\documentclass[MPS,Times1COL]{WileyNJDv5}

\articletype{Research Article}
\received{}
\revised{}
\accepted{}
\journal{}
\volume{00}
\copyyear{2026}
\startpage{1}

\usepackage{xeCJK}
\usepackage{graphicx}
\usepackage{multirow}
\usepackage{booktabs}
\usepackage{colortbl}
\usepackage{xcolor}
\usepackage{tikz}

\usepackage{amsmath,amssymb,amsfonts}
\usepackage{amsthm}
\usepackage{mathtools}
\usepackage{mathrsfs}

\usepackage{algorithm}
\usepackage{algorithmicx}
\usepackage{algpseudocode}
\usepackage{listings}

\definecolor{softblue}{RGB}{70,130,180}
\definecolor{softgreen}{RGB}{46,139,87}
\definecolor{softgray}{RGB}{100,100,100}
\definecolor{warmorange}{RGB}{255,140,0}
\definecolor{rosebrown}{RGB}{139,69,19}
\definecolor{lavender}{RGB}{150,123,182}

\newcounter{subfigletter}
\renewcommand{\thesubfigletter}{\alph{subfigletter}}
\newcommand{\resetletters}{\setcounter{subfigletter}{0}}

\newcommand{\figwithlabel}[2]{%
    \stepcounter{subfigletter}%
    \begin{tikzpicture}[baseline=(current bounding box.north)]
        \node[inner sep=0] (img) {\includegraphics[width=#2]{#1}};
        \node[
            anchor=north west,
            xshift=-6pt,
            yshift=0pt,
            font=\sffamily\bfseries\large
        ] at (img.north west) {\textbf{\thesubfigletter}};
    \end{tikzpicture}%
}

\newcommand{\insertfigrow}[2]{%
    \par\noindent
    \makebox[\textwidth][c]{%
        \foreach [count=\i from 1] \fname in {#1} {%
            \ifnum\i>1\hspace{0.04\textwidth}\fi
            \figwithlabel{\fname}{#2}%
        }%
    }%
    \par\vspace{1em}%
}

\newcommand{\figsubref}[2]{\hyperref[#1]{\ref*{#1}#2}}

\begin{document}

% \linenumbers % 自动按行编号

\title{Engram Neuron Inspired Adaptive Memorization for Lifelong Learning}
\titlemark{Engram Neuron Inspired Adaptive Memorization for Lifelong Learning}

\author[1,2]{Suiyang Jiang$^{*}$}
\author[1]{Bomin Wang$^{*}$}
\author[1]{Hangqi Zhou$^{*}$}
\author[1,2]{Jianfeng Feng}
\author[1,2]{Xiahai Zhuang}

\authormark{JIANG \textsc{et al.}}

\address[1]{\orgdiv{School of Data Science},
\orgname{Fudan University},
\orgaddress{\street{220 Handan Road},
\city{Shanghai}, \postcode{200433}, \country{China}}}

\address[2]{\orgdiv{Institute of Science and Technology for Brain-Inspired Intelligence},
\orgname{Fudan University},
\orgaddress{\street{220 Handan Road},
\city{Shanghai}, \postcode{200433}, \country{China}}}


\abstract[Abstract]{%

Lifelong learning enables artificial systems to adapt to changing environments.
To achieve this, an effective artificial learner should dynamically reorganize past and new experiences, maintaining stability while enabling flexible knowledge integration.
However, existing efforts primarily mitigate catastrophic forgetting, lacking the ability to adaptively update memory akin to biological systems.
Here, by integrating the full memory life cycle of engram neurons into artificial learners, we propose a framework that constructs a structured memory from gradient representations, which regulates parameter updates and guides memory management.
Through comprehensive evaluations on four benchmarks, our approach shows promising lifelong learning capability, maintaining stability over long task sequences and supporting flexible adaptation.
Beyond the performance gains, our approach offers a computational perspective on robust and flexible lifelong learning, echoing mechanisms observed in biological memory systems.
}


\maketitle

\footnotetext[1]{These authors contributed equally to this work.}

\section{Introduction}\label{sec1}
%\textcolor{rosebrown}{\textbf{(The introduction of LL and the limitations of current arts)}}

\begin{figure}[htbp]
    \resetletters % 每个 figure 开头重置编号
    \centering
    
    \insertfigrow{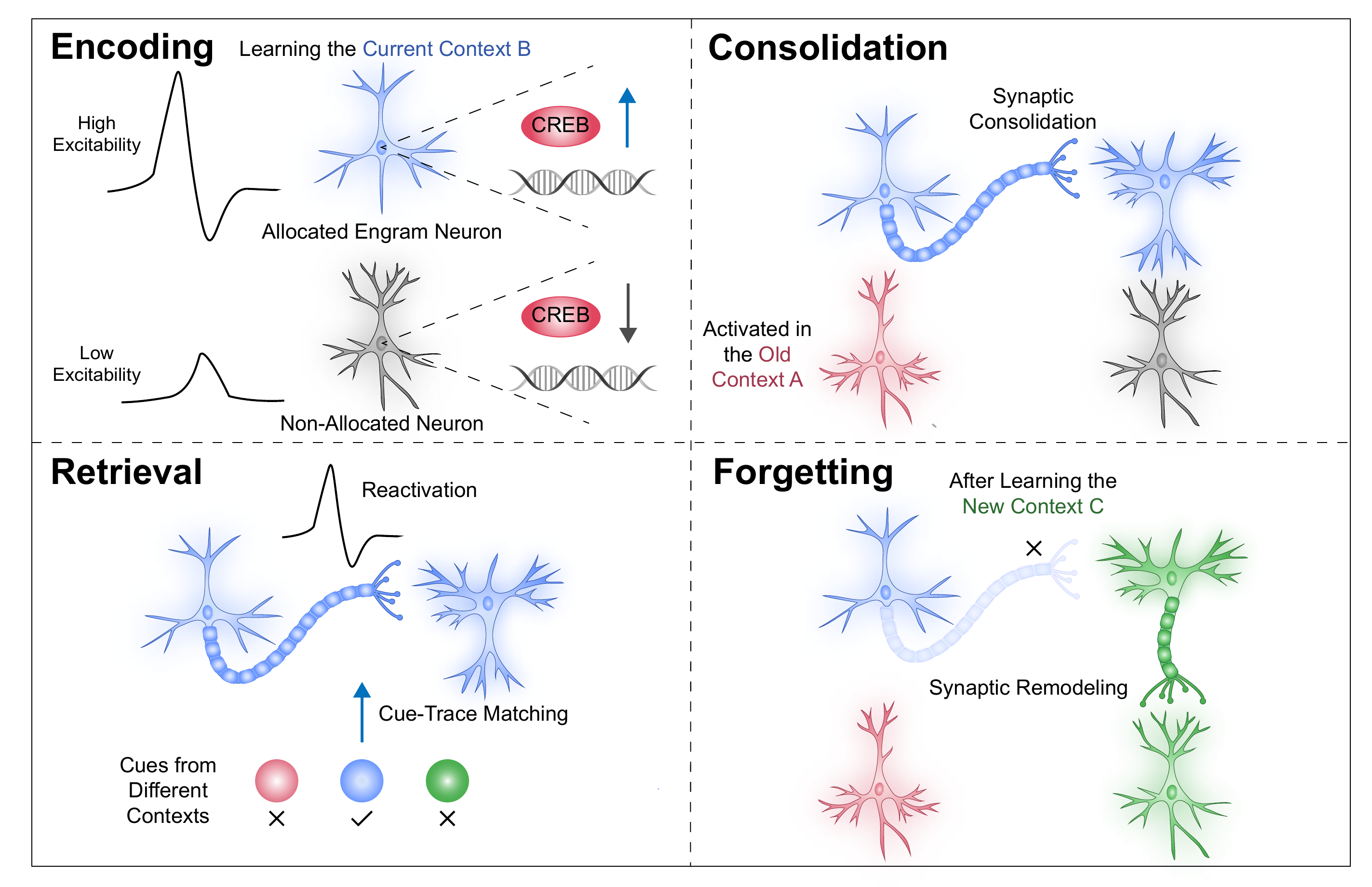}{0.8\textwidth}
    
    \insertfigrow{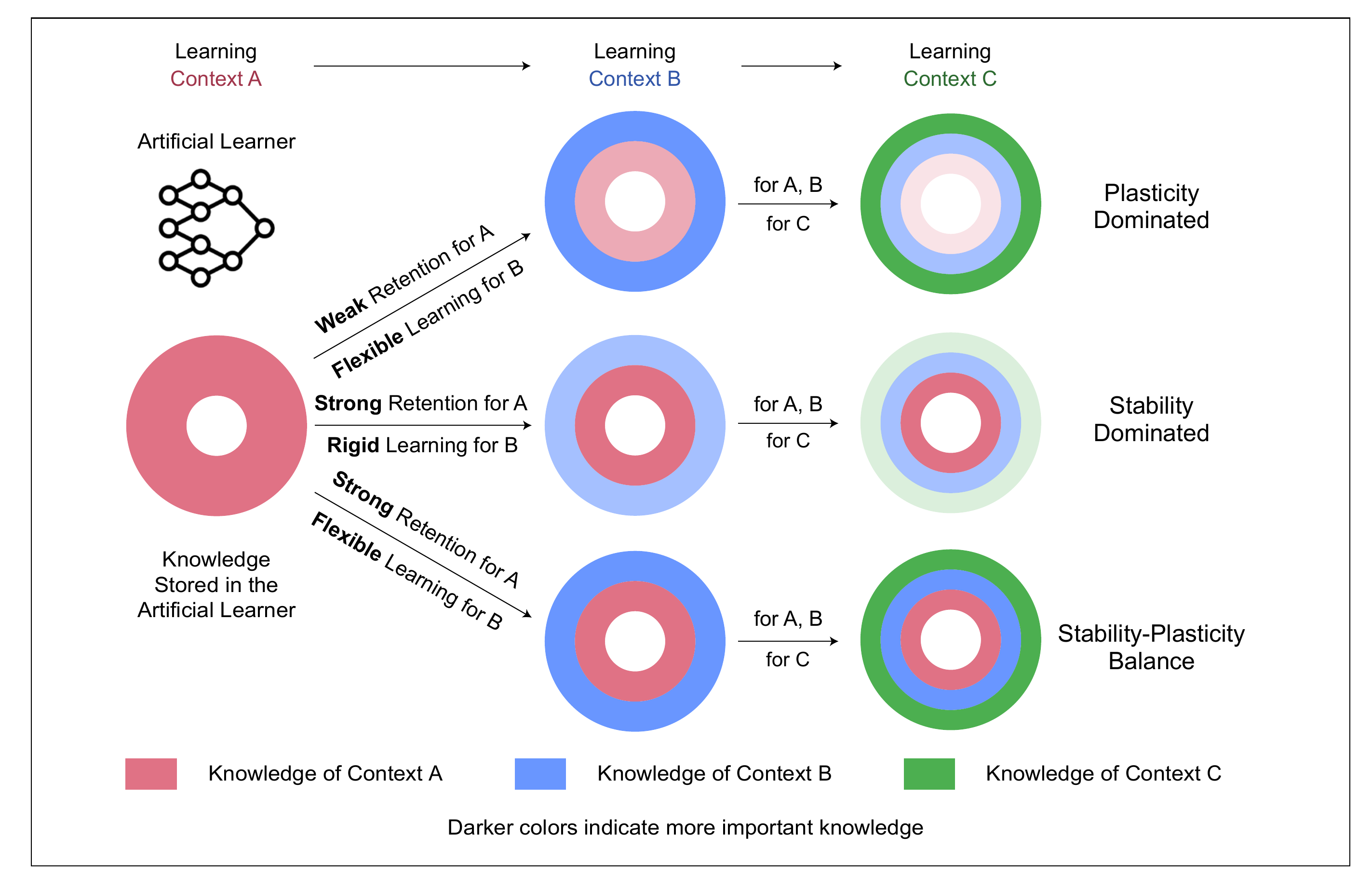}{0.8\textwidth}
    
    \caption{\textbf{Illustration of the problem in lifelong learning (LL) and how biological intelligence solves this problem:} (a) Biological memory adaptively balances old and new contexts through the lifecycle of engrams. Encoding involves CREB-regulated excitability that biases the allocation of neurons into sparse memory engrams. Synaptic plasticity underlies both consolidation and forgetting by stabilizing relevant connections and weakening or reorganizing others. Retrieval follows the encoding specificity principle, whereby cues overlapping with the original memory encoding trace preferentially reactivate the corresponding synaptic connection. CREB: cAMP response element-binding protein. (b) Plasticity dominated lifelong learning methods face severe catastrophic forgetting of previous contexts, whereas stability dominated learning restricts updating of new knowledge for new contexts. By contrast, stability-plasticity balanced LL methods adapt well to new knowledge and consolidate important prior knowledge.}
    \label{fig:introduction}
\end{figure}

%主题句：直接介绍CL的重要性——像自然生物一般持续学习
Learning underpins the ability of intelligent systems to adapt to dynamic environments \cite{1966theory, 2001learning}. The biological brain is able to incrementally acquire and update knowledge throughout its lifetime \cite{1982learning, biological_underpin}. For artificial intelligence (AI) systems aiming to adapt successfully to real-world applications, it is desirable to learn continually, similar to the biological intelligence systems. 
%CL的现状
Extensive research on lifelong learning (LL) has focused on mitigating catastrophic forgetting, a phenomenon in which learning new tasks often impairs performance on previously acquired ones \cite{1999catastrophic}. Widely adopted strategies include replaying distributions of old data\cite{2017gem, 2019er}, constructing task-specific parameters \cite{2016pnn, 2018dan}, and constraining the optimization process \cite{2017si, 2017ewc, mas}. 
However, it is still challenging to prevent forgetting and sustain learning ability simultaneously. The former is meant to enhance model stability, and the latter is meant to improve model plasticity. Unlike biological systems that benefit from an adaptive memorization mechanism, AI systems still face the challenge in maintaining the balance between them. This difficulty originates from the inherent trade-off between retaining previous knowledge and acquiring new under limited storage capacity.
Specifically, current LL methods prioritise maintenance of past knowledge but heavily neglect the learning plasticity of new tasks \cite{2024lossplasticity, 2024beyond}.

By contrast, biological intelligence can review previous knowledge from memories and effectively adapt to dynamic environments \cite{bio-advantage}. It has been observed that they can extract stable information from few samples through complex yet systematically regulated memory processes \cite{bio-memory}. 
Recent advances in neuroscience have revealed that memory is encoded and stored within specific subsets of neurons, commonly known as engram neurons. Importantly, the memory cycle is typically divided into four sequential stages, namely encoding, consolidation, retrieval and forgetting \cite{engram_four_stages_0}.

%\textcolor{warmorange}{\textbf{(Engram neuron mechanisms)}}

%Building on this memory-centric perspective, we now detail the four-stage life cycle of engram neurons. 
Fig.\figsubref{fig:introduction}{a} illuminates the four-stage life cycle of engram neuron. During encoding, a subset of neurons with high excitability, often regulated by transcription factors such as CREB (cAMP Response Element-Binding Protein), is selectively recruited to form the memory trace \cite{engram_encoding_21}. 
In the consolidation stage, those neurons with strong activation undergo long-term potentiation and synaptic stabilization, transforming short-term representations into durable long-term memories \cite{engram_consolidation_52}. 
Retrieval is governed by the encoding specificity principle, in which memory recall is successfully triggered only when external cues sufficiently match the original encoding traces \cite{engram_overall_5}. 
Finally, forgetting occurs through processes such as hippocampal neurogenesis. During this stage, newly generated neurons are integrated into existing circuits. They disrupt prior synaptic connections, and reduce the reactivation likelihood of old engram cells \cite{engram_Forgetting_128}. Therefore, this stage enables structural reallocation of memory resources.

%\textcolor{softgreen}{\textbf{(Our method summary)}}

Fig.\figsubref{fig:introduction}{b} presents the three scenarios of LL, two of which are dominated solely by stability or plasticity, thus having distorted balance. By contrast, biological intelligence typically achieves balance between the two.
Drawing inspiration from engram neurons, we propose a gradient-space-based LL method to functionally emulate each phase of the four-stage life cycle. We refer to this framework as \textbf{AdaMLearn}, short for \textbf{Ada}ptive \textbf{M}emorization for lifelong \textbf{Learn}ing. 
First, we introduce a selective sampling method to emulate the encoding stage. Specifically, AdaMLearn identifies the most representative samples from each task, feeds them into the neural network, and encodes their activations as layer-wise core memory features. This encoding scheme reduces the storage requirement and enables a compact yet comprehensive representation of task-relevant information.
Second, AdaMLearn decomposes the features to form a subspace for knowledge consolidation. This subspace is composed of gradients induced by the layer-wise features and serves as a representation of acquired knowledge. We retain only the most salient and strongly activated components, which enhance the preservation of essential knowledge and improve the capability to resist forgetting.
Third, we adopt gradient projection to simulate memory retrieval. During the training process of current task, AdaMLearn projects the updating gradients onto the aforementioned subspace and only retains the orthogonal components as the descent directions. Similar to the encoding specificity principle in neuroscience~\cite{engram_retrieval_106}, this retrieval scheme enables the model to memorise the related parameters when the current samples match previous knowledge. Therefore, it saves the neural network from catastrophic forgetting.
Finally, we propose an adaptive forgetting strategy to remove less useful memories. Specifically, AdaMLearn has an inherent mechanism to discard redundant components that hardly contribute to the overall memory utility across contexts. Together with consolidation, this forgetting process manages the storage capacity more effectively, thereby improving the adaptability of LL models.
In summary, these neuro-inspired processes constitute an integrated and interdependent memorization system that allows AdaMLearn to manage knowledge across the entire lifetime, ultimately supporting both robustness and flexibility in lifelong learning.

The proposed AdaMLearn achieves strong performance across diverse LL benchmarks, demonstrating the ability to balance stability and plasticity in complex scenarios. 
We further validate the inseparable role of the full memory cycle, highlighting how each stage contributes synergistically to LL, thereby offering insights into the mechanisms underlying real-world adaptability.

\section{Methods}\label{sec_methods}
\begin{figure}[htbp]
    \resetletters % 每个 figure 开头重置编号
    \centering
    
    \insertfigrow{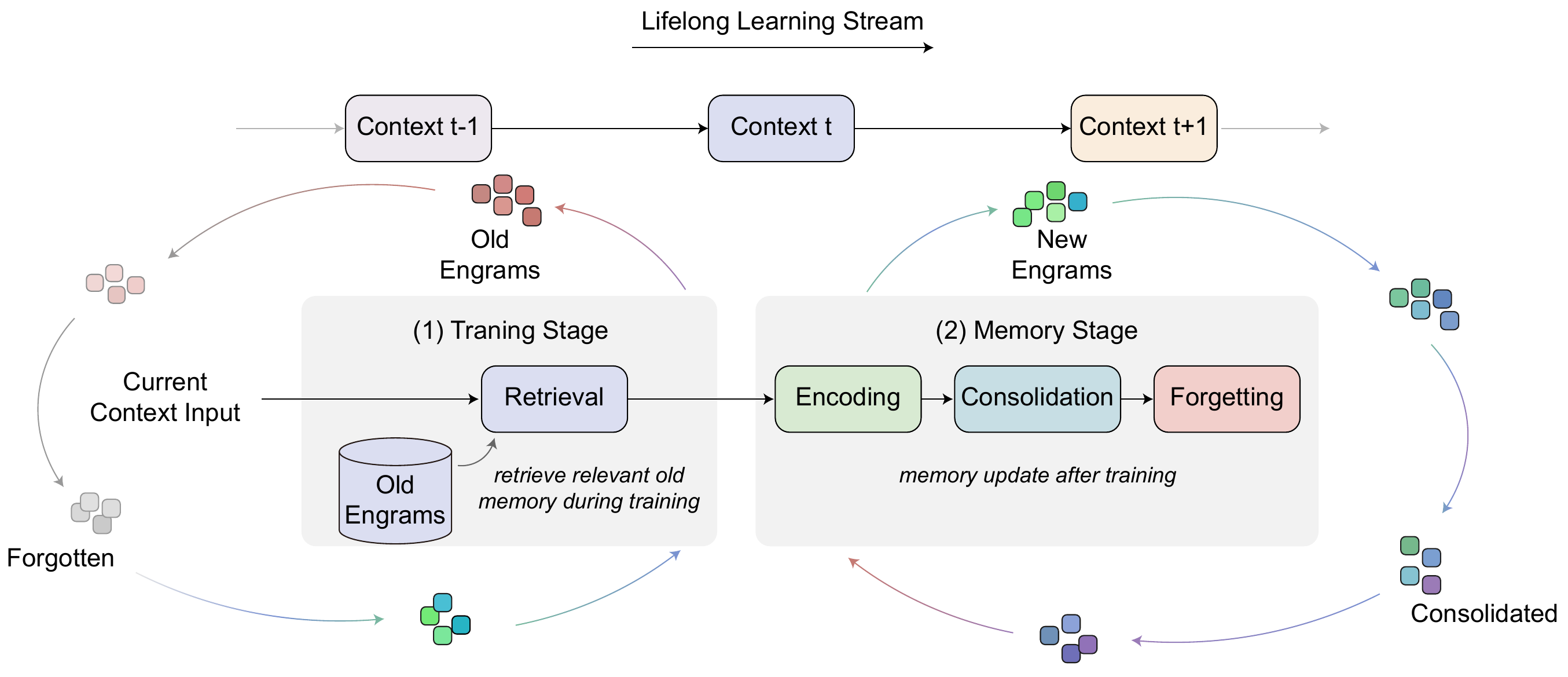}{0.6\textwidth}
    \insertfigrow{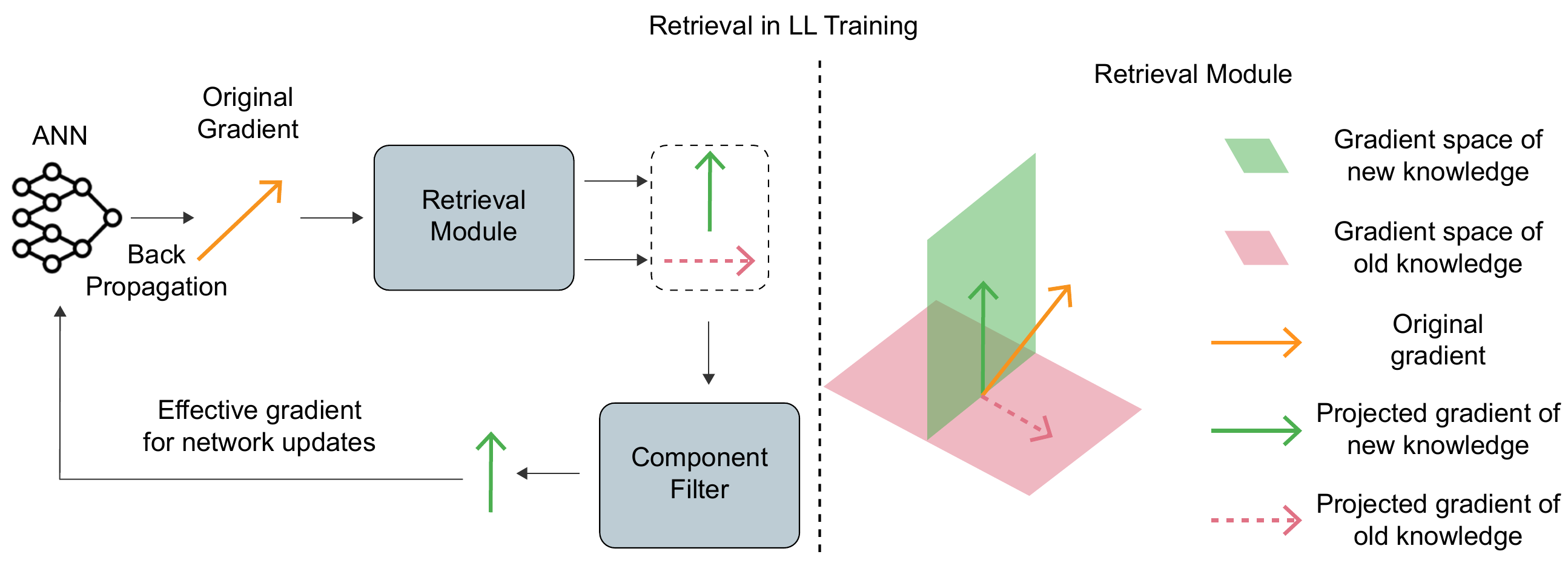}{0.6\textwidth}
    
    \insertfigrow{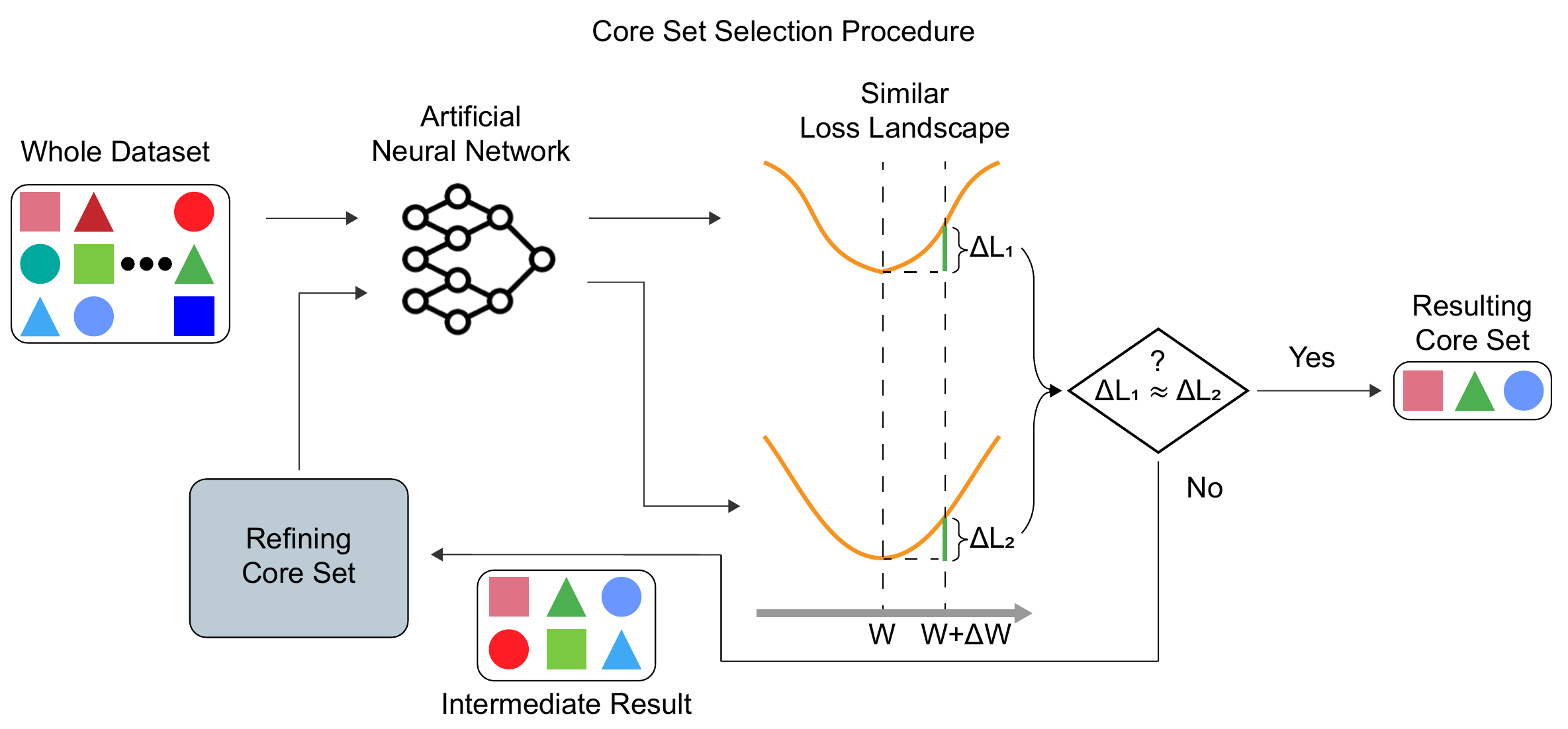}{0.6\textwidth} 
    \insertfigrow{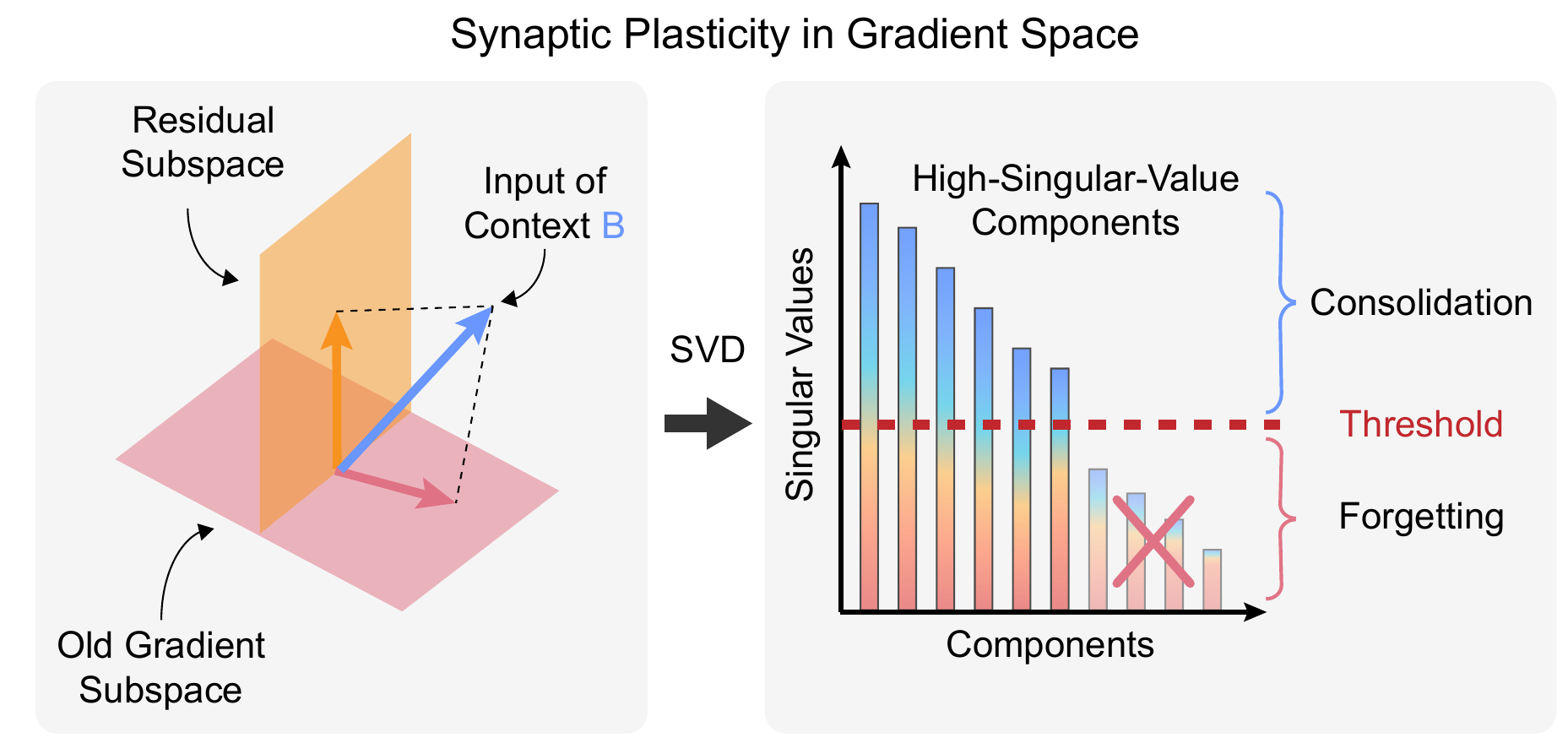}{0.6\textwidth}

    \caption{\textbf{Overview of the proposed AdaMLearn framework:} (a) Overall workflow of the proposed AdaMLearn framework across the lifelong learning stream. (b) Diagram of how the model retrieves memory during training. The initial gradient obtained from backpropagation (treated as the cue) is orthogonally projected (matching) onto the old memory (treated as the trace). The residual gradient component lies in the new memory, enabling the model to learn new knowledge without compromising previously acquired knowledge. (c) Diagram of the encoding process. we design a core set optimization procedure. In this procedure, representative samples are chosen so that the loss landscape they produce closely matches that of the full dataset. This ensures that the most important memories of the context are preserved. (d) Illustration of the proposed process in gradient space. For input data from Context B, AdaMLearn projects features onto both the old (Context A) and new (Context B) gradient subspaces, applies SVD, then consolidates high-singular-value components and forgets low-singular-value ones.}
    \label{fig:methods}
\end{figure}

\subsection{Memory Life Cycle in Lifelong Learning}
Consider an artificial neural network (ANN) trained on a sequence of tasks. During each task-incremental stage (also referred to as a task or context), only data from the current task are accessible. After sequentially learning the entire task sequence, the resulting model is then evaluated on all previously seen tasks. This setup defines the ANN as a lifelong learner (LLer). Similarly, a biological LLer will experience a series of contexts throughout its lifetime, where the knowledge acquired in each new context may overlap with, extend beyond, or even conflict with previously learned knowledge. The LLer is then expected to integrate knowledge in these contexts and maintain a competent performance throughout.

We define the memory as a gradient subspace spanned by multiple basis vectors. 
Each basis corresponds to an activated engram, and together they form a structured subspace that captures essential learning directions across contexts. 
Formally, for each layer $l$, $l = 1, \ldots, L$, the learner maintains a memory matrix $M^{\left(l\right)}$, and $M$ is referred to as the memory matrix for the entire learner:
\begin{equation}
M^{\left(l\right)} =\left [ \boldsymbol{u}_{1,l}, \boldsymbol{u}_{2,l}, \dots , \boldsymbol{u}_{n_l,l} \right ],  \quad
M=\operatorname{diag}\left \{ M^{\left(1\right)}, M^{\left(2\right)}, \dots , M^{\left(L\right)} \right \}, 
\end{equation}
where the stored gradient directions are orthogonal to each other, \textit{i.e.}, $\langle \boldsymbol{u}_{i,l},\boldsymbol{u}_{j,l}\rangle = 0$ for $i\neq j$.

The LL process unfolds sequentially in four stages, \textit{i.e.}, retrieval, encoding, consolidation, and forgetting. 
Note that for the first context, the LLer has no prior memory, thus only encoding and consolidation are applied. 
Each of the four stages modifies the memory in a distinct way.
\begin{itemize}
    \item \textbf{Retrieval.} 
    To ensure that the LLer acquires new knowledge without compromising prior capabilities, we constrain parameter updates to directions orthogonal to the memory. Specifically, the gradients are projected onto the subspace spanned by the stored memory representations from previous tasks, and the projected components are discarded during backpropagation to prevent interference with prior knowledge (Fig.\figsubref{fig:methods}{b}). This approach can be interpreted as mimicking the cue-trace matching required for successful retrieval: we retain the previous memory and enforce that learning new tasks does not overwrite prior engrams. 
    
    \item \textbf{Encoding.} To preserve informative experiences as memory, the LLer chooses a small set of samples that best represent the full dataset, referred to as core set (Fig.\figsubref{fig:methods}{c}). To quantify the importance of samples, we use the loss landscape under fixed model parameters as an indicator of neural excitability, selecting those context-specific core memories. The chosen samples are then used to generate layer-wise representation matrices $X^{\left ( l-1 \right ) }$ for memory construction.
    
    \item \textbf{Consolidation and forgetting.} Consolidation is to strengthen the distinctive features of the current context without disturbing the stored ones, and forgetting aims to free capacity for future learning by removing less useful memory. For each $X^{\left ( l-1 \right ) }$, the LLer applies singular value decomposition (SVD). As Fig.\figsubref{fig:methods}{d} illustrates, high-singular-value components from current task are integrated into the memory, selectively replacing components retained from previous tasks that are weakly activated under current task. 
\end{itemize}

Through these steps, the memory is adaptively updated, \textit{i.e.}, retrieval protects old knowledge; encoding captures key representations; consolidation strengthens important memory; and forgetting removes useless ones. This organization enables a balanced and sustainable LL process.

\subsection{Memory Retrieval During Learning}
Consider that the learner sequentially experiences learning contexts A and B, aiming to incorporate the knowledge of both tasks. When training on dataset $D_A$, the learner updates freely since there is no constraint. After training on $D_A$, the learner gets well-selected memory $M^{\left(l\right)}=\left [ \boldsymbol{u}_{1,l},\boldsymbol{u}_{2,l},\dots ,\boldsymbol{u}_{r_{A}^{\prime},l} \right ] $ of the $l$-th layer,
where the computation of $r_{A}^{\prime}$ will be described in detail later in the section \ref{section:4.4}.

When the learner experiences learning context B, only the dataset $D_B$ is available. Before taking gradient descent, the learner retrieves the memory of previous context, and projects the gradient onto vector space formed by the memory $M=\operatorname{diag}\left \{ M^{\left(1\right)}, M^{\left(2\right)}, \dots , M^{\left(L\right)} \right \}$. The learner is then updated by the gradient $\nabla_{W} \mathcal{L}$ that only retain the orthogonal parts,
\begin{equation}
% \textbf{FC Layer: }
\nabla_{W} \hat{\mathcal{L}} \coloneqq \nabla_{W}\mathcal{L} - (\nabla_{W} \mathcal{L}) M M^T.
\label{eq:R1}
\end{equation}
Appendix 1.1 presents the theoretical Analysis for layer-wise, which clarifies how the formulation above is obtained. The above memory retrieval allows the learner to learn incrementally from context B without forgetting context A.

\subsection{Engram Encoding Stage}
The learner needs to build a memory of learned knowledge after context-specific training, which emulates the encoding stage of activating engrams. As cellular excitability is the output of neurons when a learning context is taken as input, the loss also reflects the ANN’s output in response to the input data. We propose to use the loss landscape as a metric of the ANN’s response to inputs, motivated by prior evidence showing that the geometry of the loss landscape critically shapes the ANN’s ability to resist forgetting in LL~\cite{sam}. The loss landscape is described by the function $\max_{\|\epsilon\|_2 \leq \rho} \mathcal{L}(D; \theta + \epsilon)$,
where $\epsilon$ is the perturbation vector to the ANN parameter $\theta$, and $\rho$ denotes the maximum length of the perturbation vector. For the training dataset $D_A$ of context $A$, we want to find a subset $S_A$ of $D_A$, where the learner has consistent performance when taking $D_A$ or $S_A$ as input,

\begin{equation}
    \min_{S_A} \max_{\|\epsilon\|_2 \leq \rho} \frac{\mathcal{L}_{abs}(D_A, S_A; \theta + \epsilon) - \mathcal{L}_{abs}(D_A, S_A; \theta)}{\rho},
\label{eq:opt}
\end{equation}
where $\mathcal{L}_{abs}(D_A, S_A; \theta)=\left | \mathcal{L}(D_A; \theta)-\mathcal{L}(S_A; \theta) \right | $ is the difference in loss between $D_A$ and $S_A$ on the parameter $\theta $. Minimizing the loss curvature denotes similar forgetting ability between the neural network trained on $D_A$ and $S_A$. To optimize $S_A$, let $\mathbb{H}_{D} = \nabla_{\theta}^{2} \mathcal{L}(D; \theta)$ represent the Hessian matrix of $\mathcal{L}(D_A; \theta)$, and $\lambda _{1}^{D_A, S_A}$ be the maximum eigenvalue of the matrix $\mathbb{H}_{D_A,S_A}=\mathbb{H}_{D_A}-\mathbb{H}_{S_A}$, we can derive a tractable and differentiable upper bound of equation (\ref{eq:opt}) as follows~\cite{coreset},

\begin{equation}
\begin{aligned}
& \,\,\,\quad\max_{\|\epsilon\|_2 \leq \rho} \;
  \frac{\mathcal{L}_{\text{abs}}(D_A,S_A;\theta+\epsilon)
      -\mathcal{L}_{\text{abs}}(D_A,S_A;\theta)}{\rho} \\
& \leq\;
  \left\| \nabla_{\theta}\mathcal{L}(D_A;\theta)
      - \nabla_{\theta}\mathcal{L}(S_A;\theta) \right\|_2
  + \frac{1}{2}\rho \lambda_1^{T,S}
  + \max_{\|v\|_2\le1}O(\rho^2 v^3).
\end{aligned}
\label{eq:upper bound}
\end{equation}

When the ANN converges, the Hessian matrix becomes diagonally dominant. Then a diagonal approximation of the Hessian matrix can be used to reduce computational costs. By excluding higher-order terms in equation (\ref{eq:upper bound}), the resulting alternative optimization objective is then given by,
\begin{equation}
\min_{S_A} \left\| \nabla_{\theta} \mathcal{L}(D_A; \theta) - \nabla_{\theta} \mathcal{L}(S_A; \theta) \right\|_2 + \frac{1}{2} \rho \max_{k} \left| \hat{\lambda}_k^{D_A} - \hat{\lambda}_k^{S_A} \right|,
\label{eq:E3}
\end{equation}
where $\hat{\lambda}_k^{D_A}$ and $\hat{\lambda}_k^{S_A}$ are eigenvalues of diagonal Hessian matrix $\mathbb{H}_{D_A}$ and $\mathbb{H}_{S_A}$ on k-th dimension.

\subsection{Adaptive Memory Updating: Consolidation and Forgetting}
\label{section:4.4}
We now provide a detailed description of how memory updating is performed in contexts A and B. Let $n_A=\left | S_A \right | $ denotes the number of samples in $S_A$, The samples selected from $S_A$ are represented in matrix form as $X_A^{(0)}=\left [ \boldsymbol{x}_{A,1}, \boldsymbol{x}_{A,2},\dots \boldsymbol{x}_{A,n_A} \right ] $, then the learner takes $X_A^{(0)}$ as the ANN input and obtains the layer-wise input matrices $X_A^{\left(l-1\right)}$ for the $l$-th layer. \textbf{Note that all subsequent analyses in this subsection apply to an individual layer of the model. For notational clarity and brevity, we omit the superscript or subscript $l$ that would otherwise denote the layer index.}

For the consolidation of the layer-wise memory, the learner performs SVD on $X_A$,
\begin{equation}
X_A = \sum_{i=1}^{r_{A}} \sigma_{i} \boldsymbol{u}_{i} \boldsymbol{v}_{i}^T,
\label{eq:CF1}
\end{equation}
where $r_{A}$ denotes the rank of $X_A$ and $\sigma_{i}$ is the singular value with descending order($\sigma_{1}>\sigma_{2}>\dots >\sigma_{r}$). The learner then consolidates memory according to singular values: 
\begin{equation}
r_{A}^{\prime} = \arg\max_r \left( \frac{\sum_{i=1}^r \sigma_{i}^2}{\sum_{i=1}^{r_{A}} \sigma_{i}^2} \le c_A \right),
\label{eq:CF2}
\end{equation}
where $c_A\in [0,1]$ is a consolidation hyper-parameter. The learner then preserves the left singular vectors $M=\left [ \boldsymbol{u}_{1},\boldsymbol{u}_{2},\dots ,\boldsymbol{u}_{r_{A}^{\prime}} \right ] $ as memory.

Since A is the first context, only consolidation is performed, whereas subsequent context B needs to undergo both consolidation and forgetting. After completing the training of context B, the memory $M$ updates. The learner gets well-selected samples $X_B^{(0)}=\left [ \boldsymbol{x}_{B,1}, \boldsymbol{x}_{B,2},\dots \boldsymbol{x}_{B,n_B} \right ] $ and takes $X_B^{(0)}$ as the ANN input, obtaining layer-wise inputs $X_B$ for each layer.

The subsequent memory update process consists of two steps: activation of new memory and forgetting of old memory. We project $X_B$ onto the vector space spanned by the column vectors of the memory matrix $M$, obtaining the projected matrix $\hat{X}_B$ and the orthogonal part $\tilde{X}_B$, 
\begin{equation}
\hat{X}_B=M(M)^TX_B, \quad
\tilde{X}_B=X_B-\hat{X}_B.
\label{eq:CF3}
\end{equation}
Perform SVD on both parts, we can get that,
\begin{equation}
\hat{X}_B=\sum_{i=1}^{\hat{r}_B} \hat{\sigma}_{i} \boldsymbol{\hat{u}}_{i} \boldsymbol{\hat{v}}_{i}^T, \quad
\tilde{X}_B=\sum_{i=1}^{\tilde{r}_B} \tilde{\sigma}_{i} \boldsymbol{\tilde{u}}_{i} \boldsymbol{\tilde{v}}_{i}^T.
\label{eq:CF4}
\end{equation}

The projected component $\hat{X}_B$ represents the part of the learning context B that is similar to the previous context, and $\tilde{X}_B$ reflects the new knowledge unique to context B. The learner first acquires knowledge of the new context from $\tilde{X}_B$,
\begin{equation}
\tilde{r}_{B}^{\prime} = \arg\max_r \left( \frac{\sum_{i=1}^{\hat{r}_{B}} \hat{\sigma}_{i}^2+\sum_{i=1}^{r} \tilde{\sigma}_{i}^2}{\sum_{i=1}^{\hat{r}_{B}} \hat{\sigma}_{i}^2+\sum_{i=1}^{\tilde{r}_{B}} \tilde{\sigma}_{i}^2} \le c_B \right),
\label{eq:CF5}
\end{equation}
where $\tilde{r}_{B}^{\prime}$ is set to 0 when $\sum_{i=1}^{\hat{r}_{B}} \hat{\sigma}_{i}^2 \ge  c_B \left ( \sum_{i=1}^{\hat{r}_{B}} \hat{\sigma}_{i}^2+\sum_{i=1}^{\tilde{r}_{B}} \tilde{\sigma}_{i}^2 \right ) $. indicating that most of the knowledge in the current context has already been learned previously.

As new memory is activated, the old memory is then forgotten,
\begin{equation}
\hat{r}_{B}^{\prime} = \arg\max_r \left( \frac{\sum_{i=1}^r \hat{\sigma}_{i}^2}{\sum_{i=1}^{\hat{r}_{B}} \hat{\sigma}_{i}^2} \le \beta_B \right),
\label{eq:CF6}
\end{equation}
where $\beta_B\in [0,1]$ is a forgetting hyper-parameter. When $\beta_B=1$, the learner do not forget previous memory. The final memory is composed of the following,
\begin{equation}
M=\left [ \boldsymbol{\hat{u}}_{1},\dots ,\boldsymbol{\hat{u}}_{\hat{r}_{B}^{\prime}},\boldsymbol{\tilde{u}}_{1},\dots ,\boldsymbol{\tilde{u}}_{\tilde{r}_{B}^{\prime}} \right ].
\label{eq:CF7}
\end{equation}

\subsection{Implementation}

In the settings of task-incremental learning, we construct four experimental benchmarks for evaluation. The CIFAR-100 dataset~\cite{dataset-CIFAR-100} includes 100-class coloured images of size 32×32. We split it into 10 tasks according to the natural class order, with each task containing 10 classes that are disjoint from other tasks. The CIFAR-superclass divides the same 100 classes into 20 superclasses based on semantic similarity, resulting in 20 tasks, each containing 5 classes from a specific superclass. This longer sequence increases task dissimilarity and makes the overall learning process more challenging. The FIVE dataset benchmark~\cite{dataset-5-datasets} consists of 5 tasks, each being a 10-class classification problem from a different dataset, including CIFAR-10, MNIST, SVHN, Fashion-MNIST, and notMNIST. The datasets differ in image size (e.g., 28×28 and 32×32) and visual appearance, introducing significant domain shifts and increasing the difficulty of LL. The Mini-ImageNet~\cite{dataset-mini-imagenet} comprises 100 classes of images with a larger size of 84×84, split into 20 tasks. To further challenge the robustness and adaptability of the model, the class order is randomly shuffled for each run in this experiment.

\subsection{Competing Approaches}
To ensure generality in the applications, we compare with representative LL methods across different paradigms. Elastic Weight Consolidation (EWC~\cite{2017ewc}), Learning without Forgetting (LwF~\cite{2017LwF}), and Synaptic Intelligence (SI~\cite{2017si}) are classical regularization-based approaches that constrain parameter updates to preserve knowledge from previous tasks. Among replay-based methods, Averaged Gradient Episodic Memory (AGEM~\cite{2018agem}) projects gradients to avoid increasing the loss on stored samples, Experience Replay (ER~\cite{2019er}) simply replays buffered old samples alongside new data, Dark Experience Replay (DER~\cite{2020der}) combines experience replay with knowledge distillation to improve knowledge retention, its variant DERpp further incorporates distillation on logits trajectories to enhance stability. In addition, we include state-of-the-art approaches. Learning a Unified Classifier Incrementally via Rebalancing (LUCIR~\cite{2019lucir}), which mitigates class imbalance by enforcing inter-class separation and feature normalization; Bias Correction (BiC~\cite{2023bic}), which reduces prediction bias toward new classes by introducing a correction layer; and eXtended Dark Experience Replay (X-DER~\cite{2022xder}), which enhances replay-based learning with stronger distillation and dynamic memory update strategies.

All baseline methods are implemented within the Mammoth benchmark~\cite{2022xder}, and experiments are conducted with the best hyperparameter configurations. Each method is evaluated over multiple runs to ensure robust and reliable performance comparisons.
However, LUCIR and BiC could not be successfully implemented under the FIVE setting due to mismatched image sizes across different tasks. EWC failed on the Mini-ImageNet dataset due to non-convergent loss. Thus their results are not reported.

\subsection{Evaluation Metrics}
We evaluate LL performance using two standard metrics, average accuracy (AAC) and backward transfer (BWT)~\cite{GPM}:
\begin{align}
\text{AAC} \; &= \; \frac{1}{T} \sum_{i=1}^{T} A_{T,i}, \label{eq:aac} \\[6pt]
\text{BWT} \; &= \; \frac{1}{T-1} \sum_{i=1}^{T-1} \bigl( A_{T,i} - A_{i,i} \bigr). \label{eq:bwt}
\end{align}
where $T$ denotes the total number of tasks. The term $A_{T,i}$ represents the test accuracy on task $i$ after the model has sequentially learned up to task $T$, and $A_{i,i}$ indicates the test accuracy on task $i$ immediately after it was first learned.
AAC reflects the overall LL ability of the model after balancing stability and plasticity, making it the most important performance indicator. BWT quantifies stability by measuring forgetting on previous tasks when learning new ones, where more negative values indicate greater forgetting. Moreover, $A_{i,i}$ reflects model plasticity, with larger values indicating stronger capability to acquire new knowledge.

\section{Results}\label{sec2}
\subsection{Adaptive Memorization Enhances Lifelong Learning Performance}

Insights from the full memory life cycle of engram neurons suggest that stability, adaptability, and sparse neural encoding are not isolated components but tightly entangled that jointly support robust LL~\cite{engram_four_stages_0}. Our method naturally links the gradient-driven parameter updates in neural networks to memory-guided learning, reflecting how prior experiences shape future adaptation in biological systems.

%Further technical details of this procedure are provided in the Methods section.

\begin{figure}[htbp]
    \resetletters % 每个 figure 开头重置编号
    \centering
    
    \insertfigrow{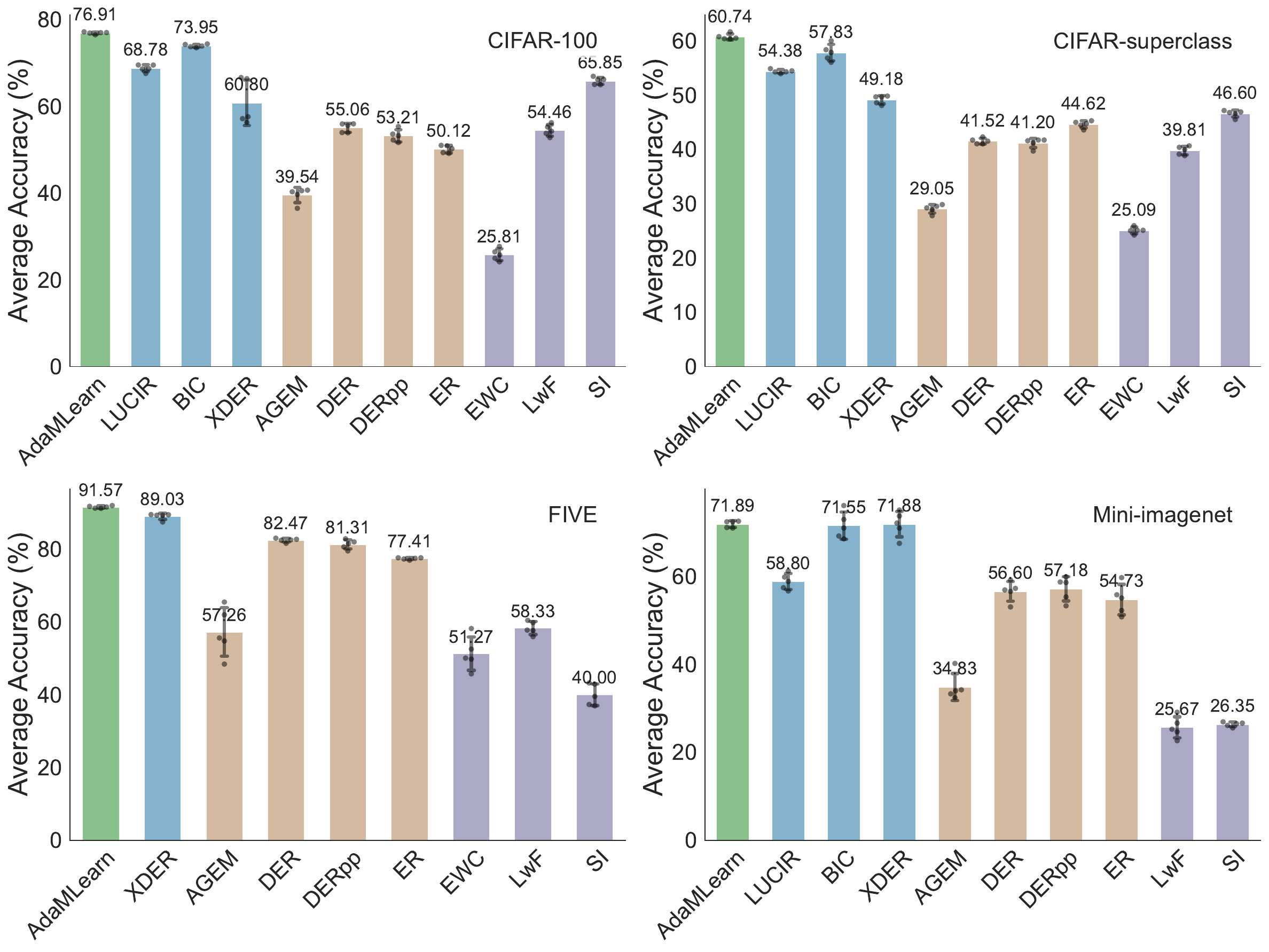}{0.8\textwidth}
    
    \insertfigrow{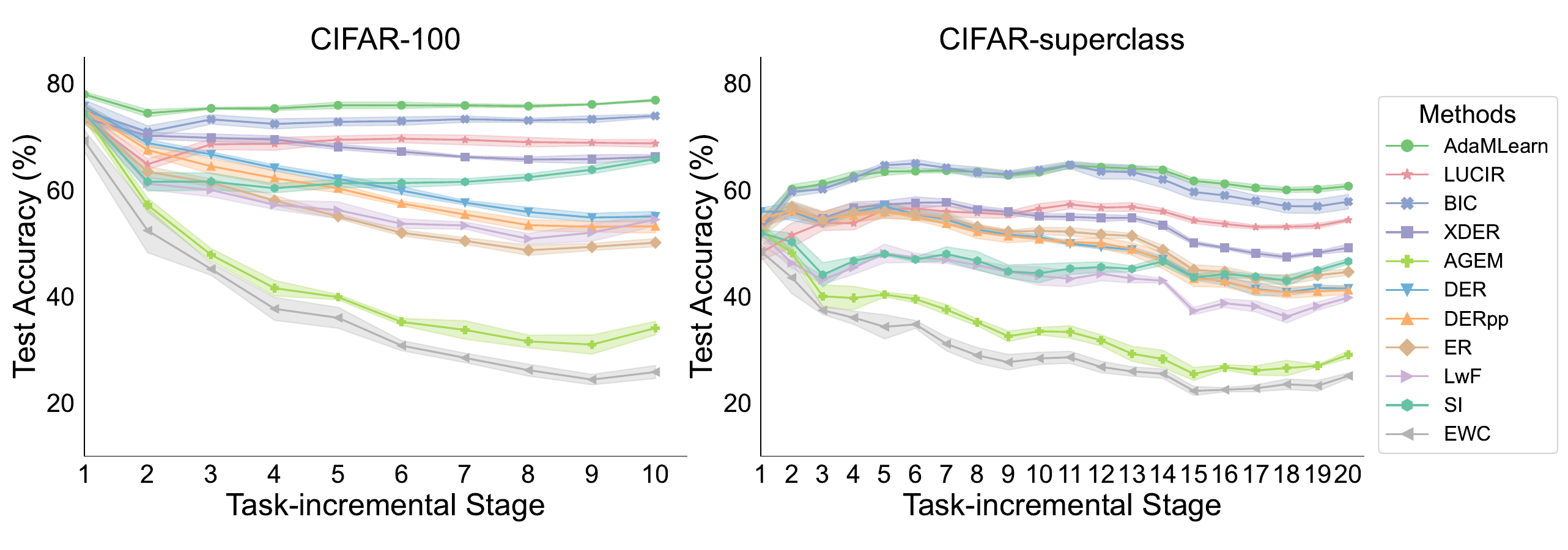}{0.8\textwidth}
    
    \insertfigrow{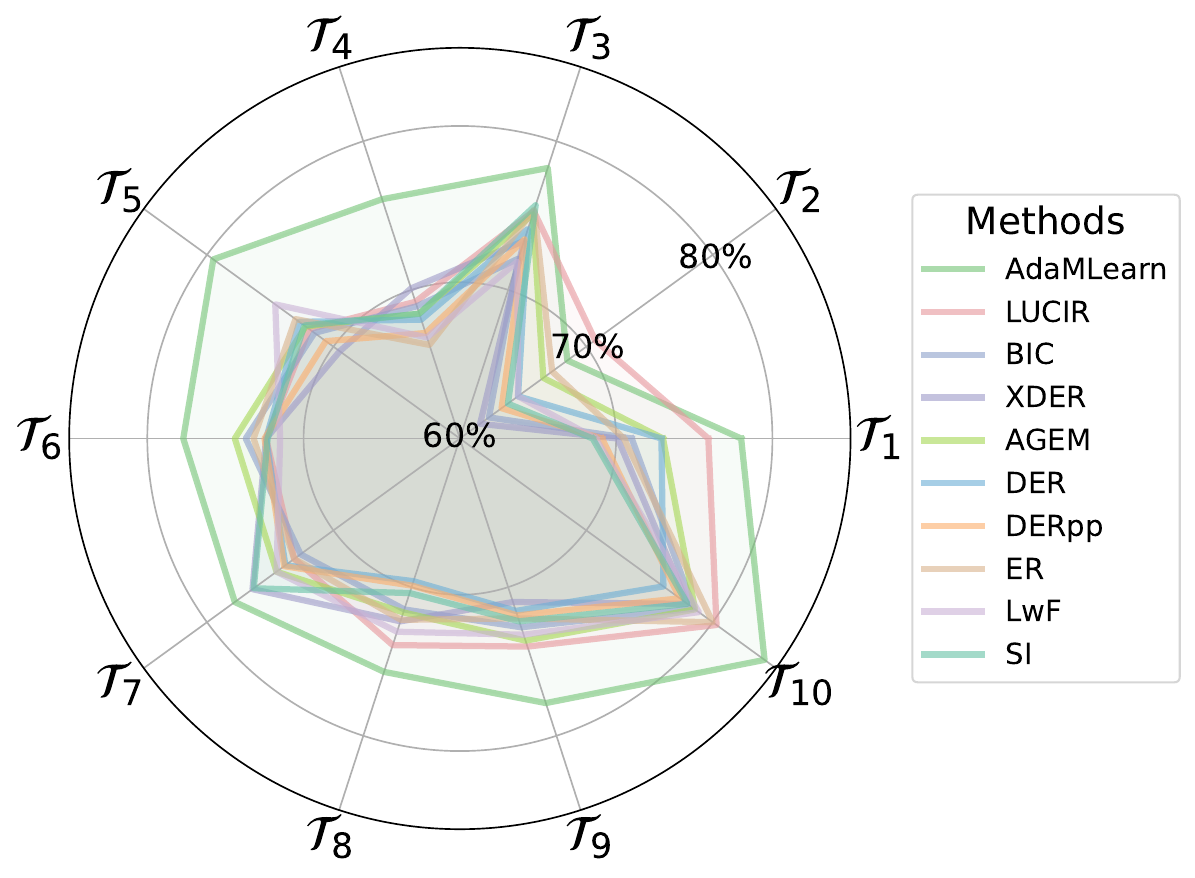, 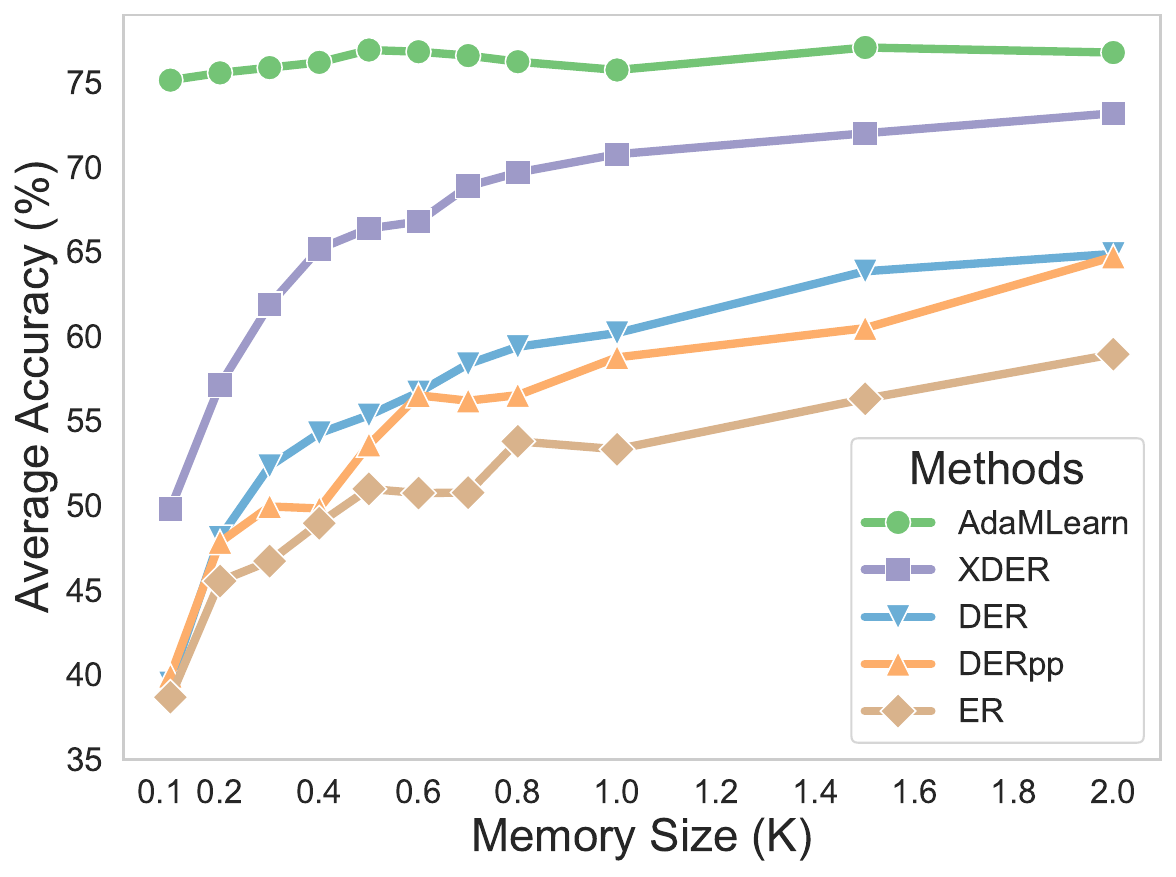}{0.34\textwidth}
    
    \caption{\textbf{Overall performance in visual classification tasks:} (a) Lifelong learning performance is evaluated from multiple perspectives, including sensitivity to task ordering, cross-domain transferability, and robustness to higher-resolution inputs. Colors represent the category of different LL methods. All results are averaged over five runs with different random seeds. (b) The learner’s performance on all previously learned tasks at each stage of the full task sequence is reported, averaged over five runs. (c) Plasticity is assessed by measuring the test accuracy on a new task immediately after its initial learning, with experiments conducted on CIFAR-100. (d) Comparison with other replay-based lifelong learning methods shows that our approach maintains strong performance across a wide range of buffer size settings. Notably, even when the number of available samples is small, it does not suffer from catastrophic forgetting.}
    \label{fig:results_1_overall}
\end{figure}

We evaluated the overall performance of our approach under the task-incremental learning setting on four image classification benchmarks and compared it with both state-of-the-art and classical LL methods. Specifically, CIFAR-100 and CIFAR-superclass represent two different context ordering strategies on the CIFAR-100 dataset; the FIVE dataset assesses the ability to transfer across heterogeneous domains; and Mini-ImageNet introduces higher-resolution images with increased visual diversity and semantic complexity, posing a more challenging and realistic lifelong learning scenario. As Fig.\figsubref{fig:results_1_overall}{a} illustrates, despite the diverse settings, AdaMLearn consistently achieves state-of-the-art performance across all benchmarks. Furthermore, AdaMLearn demonstrates robust performance across different contexts (Fig.\figsubref{fig:results_1_overall}{b}), as well as strong plasticity in adapting to new tasks (Fig.\figsubref{fig:results_1_overall}{c}). Finally, Fig.\figsubref{fig:results_1_overall}{d} illustrates the ability of AdaMLearn to accurately encode the overall memory of a context using only a core set with small size.
%Compared to typical replay-based lifelong learning methods, EN maintains competitive accuracy across a wide range of buffer sizes (Fig.~\ref{fig:results_1_overall_d}), showing only marginal degradation even when memory resources are severely limited.

\subsection{Consolidation and Forgetting Improves Adaptability}

\begin{figure}[htbp]
    \resetletters % 每个 figure 开头重置编号
    \centering

    \insertfigrow{img/results/2_consolidation_forgetting/synaptic_plasticity}{0.6\textwidth}
    \insertfigrow{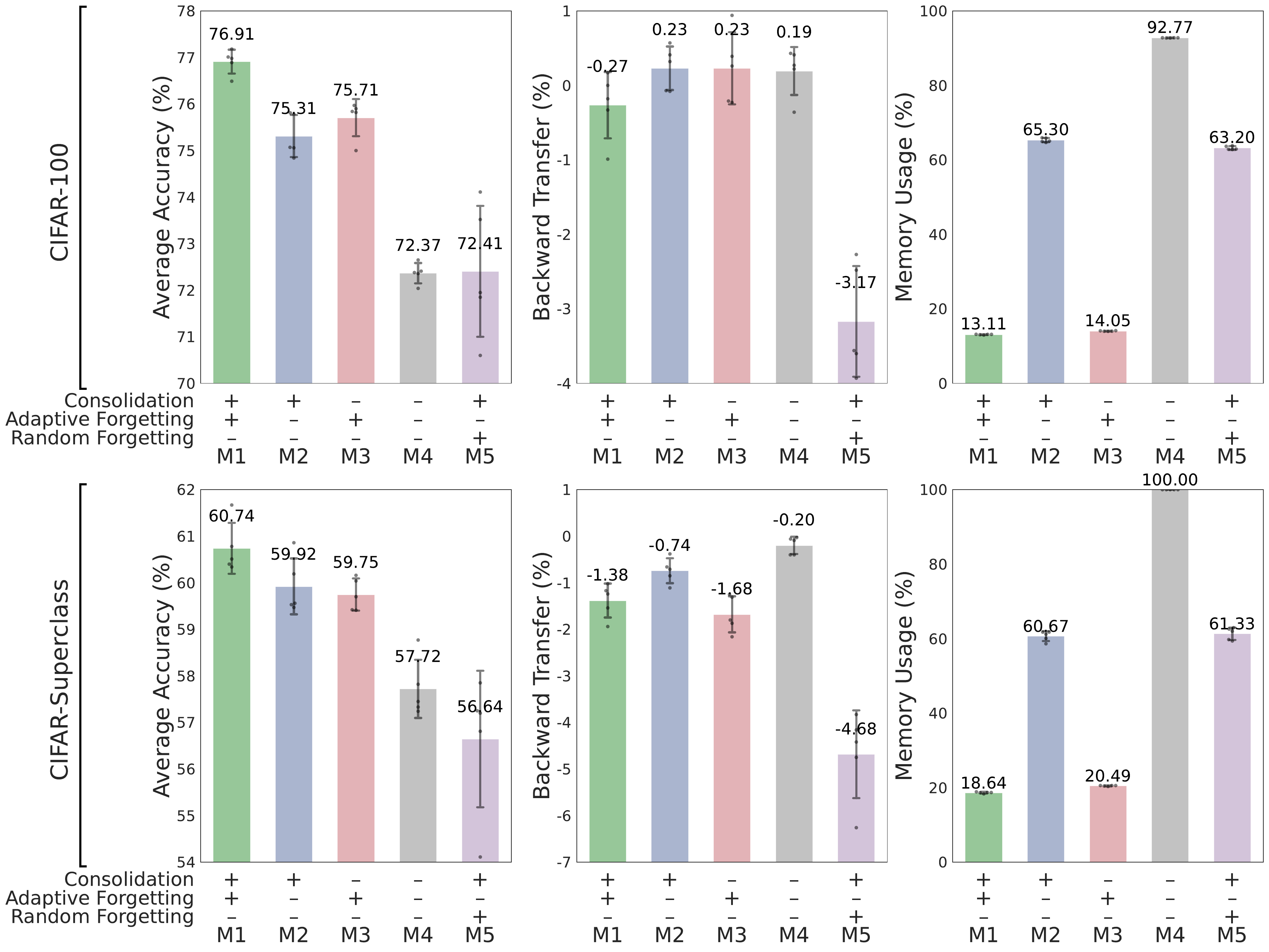}{0.6\textwidth}
    \insertfigrow{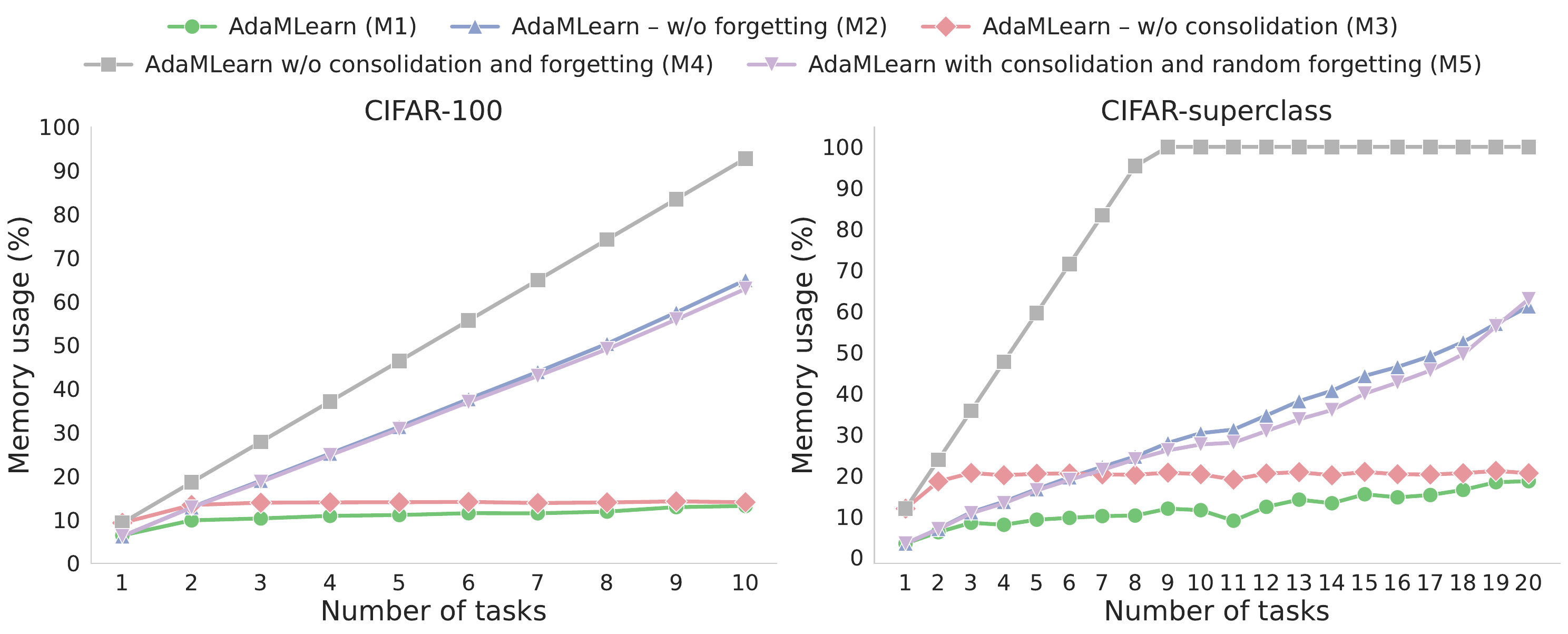}{0.6\textwidth}
    
    \caption{\textbf{Achieving lifelong adaptive memorization within the framework of synaptic plasticity:} (a) Integration of consolidation and forgetting within the neural synaptic plasticity. The color and intensity of nodes reflect the activation level of neurons. Learning context B introduces new potential synapses that partially weaken previously consolidated connections from context A. (b) Ablation experiments evaluate average accuracy and backward transfer. For AdaMLearn (M1), although synaptic remodeling increases the degree of forgetting, it leads to an overall improvement in average accuracy. All results are averaged over five runs. (c) Effects of consolidation and forgetting on memory efficiency. Without consolidation and forgetting (M4), memory grows rapidly and eventually saturates the budget. For AdaMLearn (M1), memory usage remains low and stable across tasks, adaptively converging to compact memory budgets that reflect efficient and context-aware resource allocation in CIFAR scenarios. The memory usage is computed at the level of the final linear hidden layer of the ANN. A value of 100\% indicates that subsequent learning does not alter the parameters of this layer.}
    \label{fig:results_2_consolidation_forgetting}
\end{figure}

% 巩固和遗忘是突触可塑性的两个方面。突触可塑性是生物适应环境的神经基础。
To flexibly acquire new knowledge without disrupting previously learned information, biological systems rely on synaptic plasticity, and we divide it into two stages: consolidation and forgetting (Fig.\figsubref{fig:results_2_consolidation_forgetting}{a}). In the hippocampus, changes in engram depend on neurogenesis~\cite{engram_consolidation_nature2025}. Newborn dentate granule cells (DGCs) are generated in the subgranular zone of the hippocampus and later form input and output synapses in DG and CA3. They integrate into the existing DG-CA3 circuit and compete for synaptic input and output with mature neurons. These new neurons form novel connections that may coexist with or replace old ones, resulting in circuit remodeling~\cite{engram_Forgetting_11}. Due to their higher intrinsic excitability, new DGCs facilitate the encoding of recent experiences while disrupting the reactivation probability of pre-existing engram cells~\cite{engram_consolidation_52}. This remodeling reflects a form of structural forgetting, enabling adaptive memory allocation through representational rewiring. 

Inspired by this biological process, we perform memory consolidation and forgetting within the representational gradient subspace. Results in Fig.\figsubref{fig:results_2_consolidation_forgetting}{b} demonstrate that these two processes reveal a synergistic effect. Although forgetting involves the removal of outdated knowledge, it can improve overall performance. This suggests that consolidation and forgetting are tightly coupled aspects of synaptic plasticity. Notably, we compare this mechanism with a variant employing random forgetting, where bases in the memory matrix are removed uniformly at random according to a fixed proportion. By contrast, our activity-dependent strategy, guided by the “use it or lose it” principle, yields more effective results. On CIFAR100~\cite{dataset-CIFAR-100}, the average accuracy of the proposed method reaches $76.91\%$, $4.50\%$ higher than that of the variant of random forgetting (M5). Similarly, on CIFAR Superclass, the performance increases from $56.64\%$ at the variant to $60.74\%$. 

Furthermore, in Fig.\figsubref{fig:results_2_consolidation_forgetting}{c} we track the evolution of memory capacity across tasks. One can see that on CIFAR Superclass, the AdaMLearn variant with neither consolidation nor forgetting (M4) quickly reaches its memory budget after the 9th task, indicating severe inefficiency in memory usage. By contrast, the full AdaMLearn (M1) consumes less than one-fifth memory storage at the end of the whole training period. It is worth noting that the variant with random forgetting combined with consolidation (M5) exhibits memory usage comparable to the consolidation-only setting, indicating that naive forgetting does not provide additional benefits. These results underscore the importance of the proposed adaptive forgetting mechanism, which enables the lifelong learner to dynamically allocate memory resources according to task demands and maintain sufficient capacity for future adaptation.

\subsection{Memory Encoding Increases Efficiency}

\begin{figure}[htbp]
    \resetletters % 每个 figure 开头重置编号
    \centering
    \insertfigrow{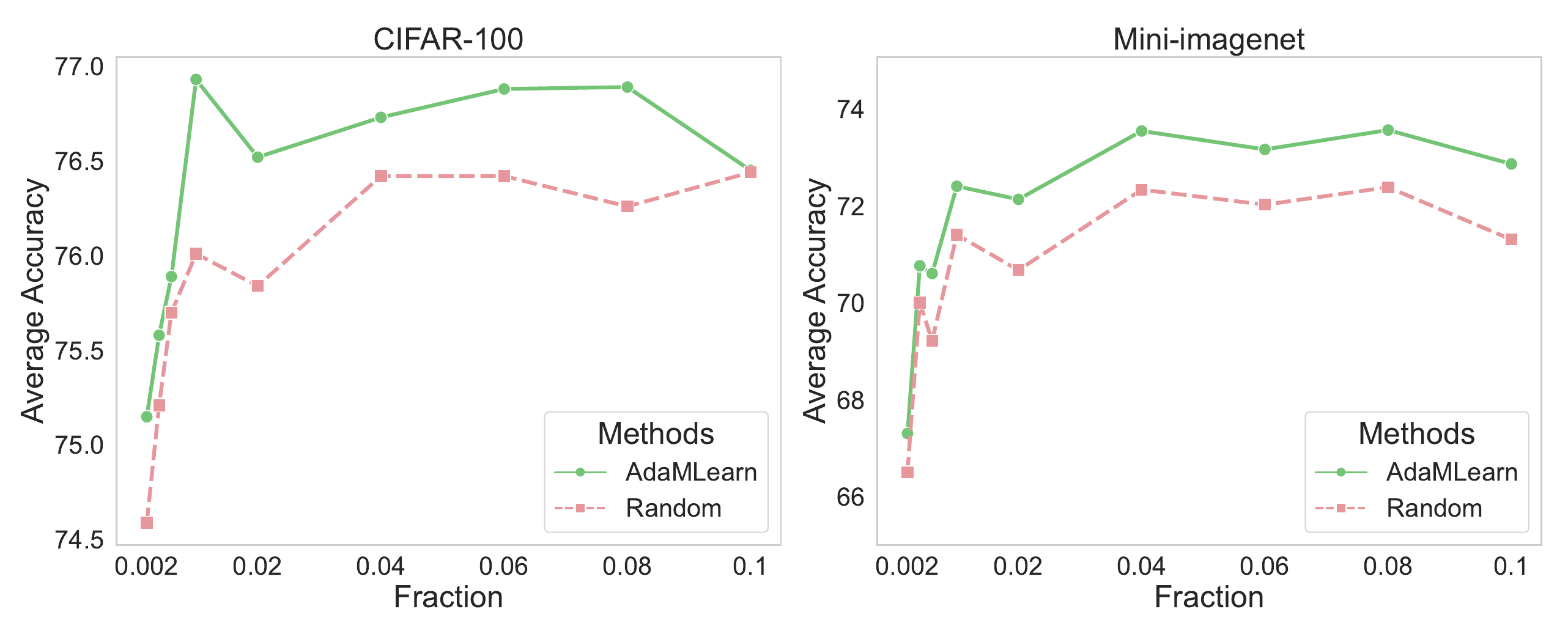}{0.7\textwidth}
    \insertfigrow{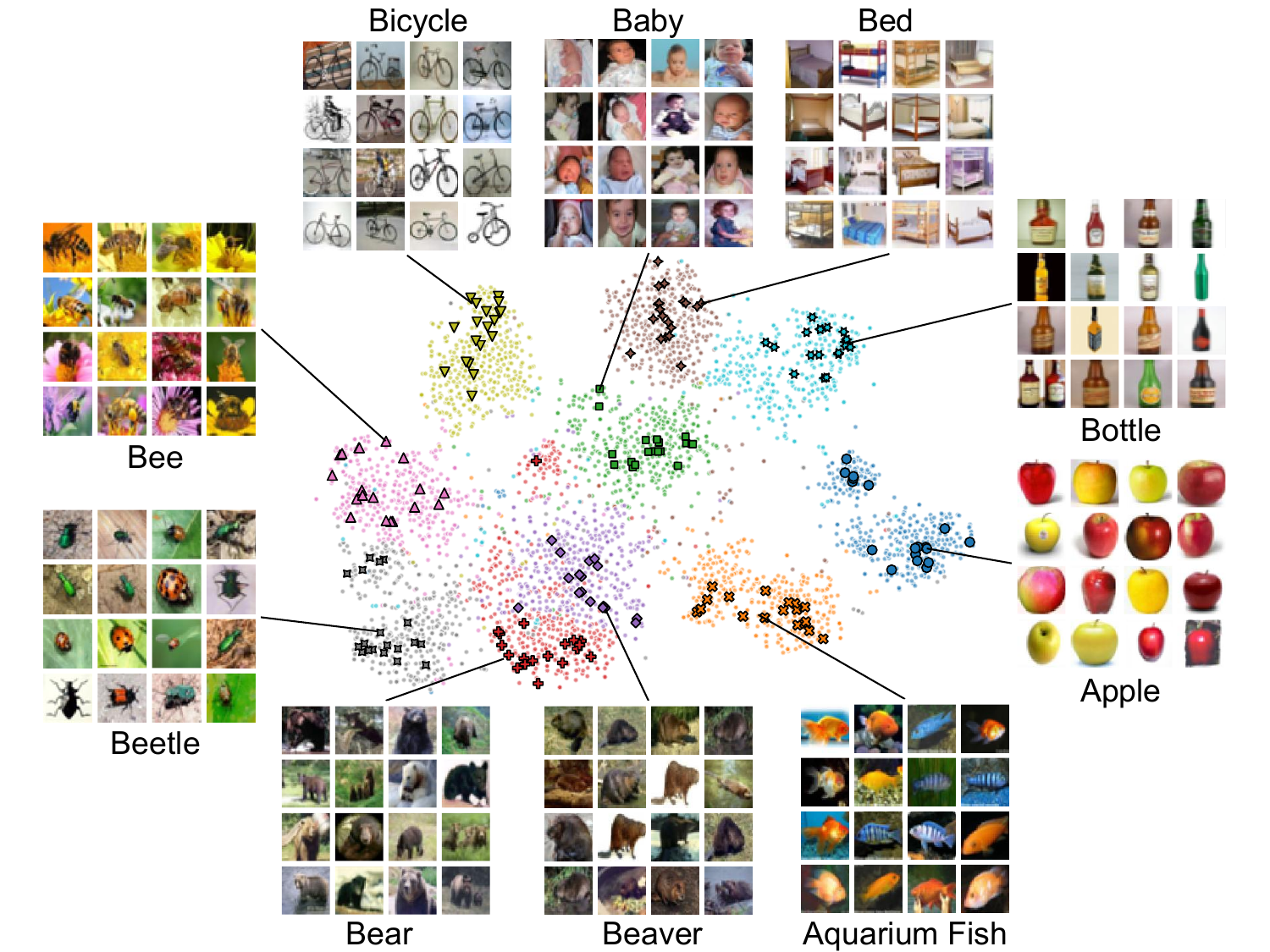}{0.7\textwidth}
    
    \caption{\textbf{Implementing the selective encoding of contextual memory:} (a) Effects of the core set selection procedure. Varying the fraction of core set selection demonstrates that our encoding process yields substantial benefits across a wide range of selection ratios. The baseline (Random) means selecting samples randomly. (b) The t-SNE plot of the CIFAR-100 dataset illustrates the representative samples chosen for each class after core set optimization. Samples selected as core set are marked with distinctive shapes, whereas the unselected ones appear as small circular points.}
    \label{fig:results_3_coreset}
\end{figure}

Previous studies have shown that only a subset of eligible neurons within a neural network is recruited to encode a particular experience~\cite{engram_encoding_34}. This allocation process tends to favor neurons with higher intrinsic excitability over those with reduced excitability. The transcription factor CREB plays a key role in regulating this excitability~\cite{engram_encoding_21, engram_encoding_26}. Experimental evidence has demonstrated that neurons overexpressing CREB are more likely to fire action potentials in response to identical stimuli, compared to their neighbors. These neurons also exhibit reduced afterhyperpolarization potentials (AHP), a property associated with decreased $K^+$ channel conductance~\cite{engram_encoding_34}. As a result, increasing CREB levels in specific neurons significantly raises their probability of being incorporated into an engram, whereas lowering CREB expression diminishes this likelihood. This phenomenon is referred to as memory allocation.

Inspired by the selective recruitment of highly excitable neurons during biological memory allocation, AdaMLearn implements an encoding strategy that preferentially selects a subset of representative experiences to construct memory. Fig.\figsubref{fig:results_3_coreset}{a} demonstrates the effectiveness of these core memories in supporting LL performance. Among different fractions, our proposed encoding process outperforms the baseline (Random) in terms of average accuracy. In the CIFAR-100 experiment, at a fraction of 0.10, the enlarged sample pool renders random selection comparably effective, leading to performance similar to that of the proposed core-set strategy. Fig.\figsubref{fig:results_3_coreset} {b} illustrates the resulting core set of the encoding process which selects a distinct and representative set of samples from each class. For instance, the selected apple samples are distinguished from other classes and come in diverse colors and shapes. 

\subsection{Memory Retrieval Maintains Stability}

\begin{figure}[htbp]
    \resetletters % 每个 figure 开头重置编号
    \centering
    \insertfigrow{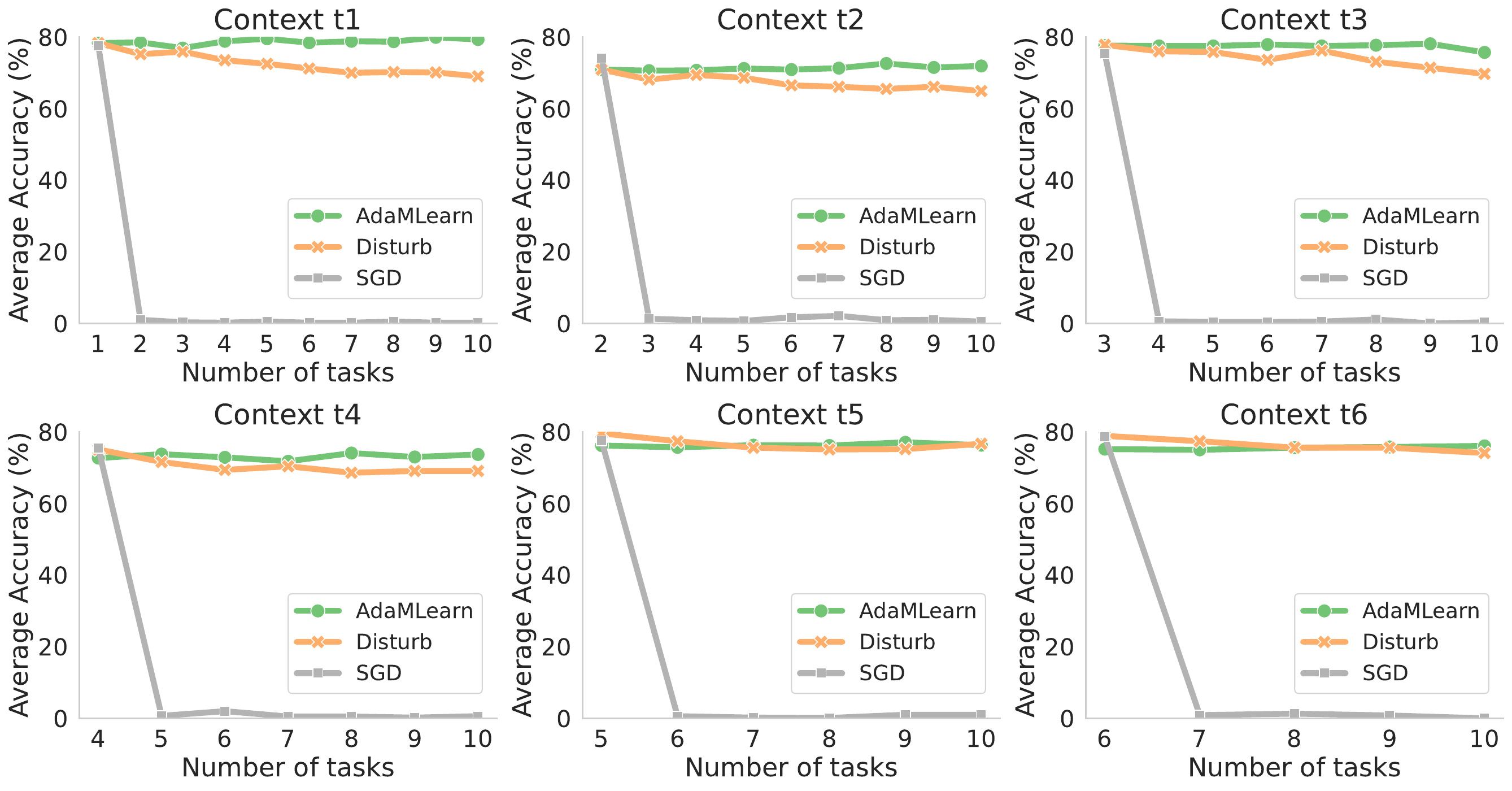}{0.95\textwidth}
    
    \caption{\textbf{Applying retrieval during LL:} (a) Effects of the retrieval process on preserving prior knowledge. Without any constraint (SGD), the model rapidly forgets previously acquired knowledge when learning a new task. When perturbations are introduced during retrieval (Disturb), causing the parameter update direction to deviate from being strictly orthogonal to the old memory, the stability degrades.}
    \label{fig:results_4_retrieval}
\end{figure}

In neuroscience, memory retrieval is governed by the Encoding Specificity Principle, which states that successful recall depends on the overlap between the retrieval cues and the present memory conditions~\cite{engram_overall_5}. This implies that only specific cues that match the memory trace of the original engram can effectively trigger its reactivation. Importantly, retrieval often fails not because the memory is lost, but because the cues are insufficient to access the original engram~\cite{engram_overall_5}. This highlights memory retrieval as a content-dependent process rather than a global reactivation of stored information. 

Inspired by this principle, we introduce this cue-aligned retrieval mechanism based on gradients to resist forgetting. Fig.\figsubref{fig:results_4_retrieval}{a} compares the average accuracy across sequential tasks. AdaMLearn maintains stable performance as tasks accumulate, showing almost no degradation. By contrast, the variant (Disturb) suffers moderate accuracy decay since cue-trace matching is disrupted. Vanilla SGD rapidly collapses to near-zero accuracy after only one task. These results illustrate that preserving cue-trace consistency effectively prevents memory interference, whereas unconstrained learning inevitably triggers catastrophic forgetting.

\section{Discussion}\label{sec3}

The ability to learn continually is essential for building AI systems that remain useful in dynamic, real-world environments such as healthcare, autonomous systems, and embodied AI.
By incorporating the full memory cycle of engram neurons, the proposed AdaMLearn enables the AI model to efficiently integrate new information and preserve previously acquired knowledge, with minimal storage requirements. These characteristics make it highly suitable for practical applications operating in evolving contexts. Furthermore, by eliminating the need for exhaustive retraining on past data, this paradigm reduces computational overhead, suggesting a more resource-efficient direction for scalable lifelong learning systems.

In biological systems, memory achieves lifelong adaptability by dynamically balancing stability and flexibility to cope with changing environments, a principle that our work brings into the context of lifelong learning. An important line of research in neuroscience has studied memory processes at the cellular level through the engram neuron framework, which describes memory as a four-stage life cycle and provides a clear biological basis for concepts that are otherwise abstract \cite{engram_four_stages_0}. Our gradient-space-based framework integrates these mechanisms into the learning architecture. Selective encoding focuses on core representations that best capture the essence of each context. Consolidation and forgetting together implement an adaptive 'use-it-or-lose-it' strategy, allowing the memory to change according to relevance. Retrieval, guided by cue-trace alignment, controls gradient updates by replaying knowledge within the subspace of previous memories. This approach not only offers a biologically consistent explanation for achieving lifelong learning in artificial systems, but also suggests practical directions for how structured memory updates and representational adjustments can support real-world adaptability.

Artificial and biological intelligence share the common goal of adapting and thriving in dynamic environments, and the two fields have a strong potential to learn from and advance each other. By incorporating these interdependent memory mechanisms into artificial learners, our approach offers a biologically informed perspective on improving learning plasticity and efficiency. Future research could extend this framework to broader modalities, incorporate additional principles underlying natural intelligence, and explore synergies with neuromorphic architectures. Such efforts may progressively evolve current AI systems towards more adaptive, energy-efficient, and resilient forms of intelligence.

\section*{Data Availability}
All benchmark datasets used in this paper are publicly available, including CIFAR-100~\cite{dataset-CIFAR-100} (\url{https://www.cs.toronto.edu/~kriz/cifar.html}), FIVE dataset~\cite{dataset-5-datasets} (\url{https://github.com/facebookresearch/Adversarial-Continual-Learning}), and Mini-ImageNet~\cite{dataset-mini-imagenet} (\url{https://huggingface.co/datasets/timm/mini-imagenet}).

\section*{Code Availability}
The implementation code is available via \url{https://github.com/xhgb/AdaMLearn}.

% \section*{Acknowledgment}
% This work was funded by the National Natural Science Foundation of China (grant No. 62372115), Shanghai Municipal Education Commission-Artificial Intelligence Initiative to Promote Research Paradigm Reform and Empower Disciplinary Advancement Plan (grant no. 24KXZNA13), and Shanghai Oriental Elite Talent (YingCai) Program. We thank Yibo Gao, Sihan Wang, Hongbin Han, Yi Tang and Xiao Xiao for helpful discussions.

% \section*{Author contributions}
% S.J., B.W., H.Z. and X.Z. conceived the project. S.J. developed and trained the computational model, and performed the theoretical analysis. S.J., B.W. and H.Z. performed all experiments and wrote the paper. S.J., B.W., H.Z., J.F. and X.Z. revised the paper. X.Z supervised the project.

\section*{Competing interests}
The authors declare no competing interests.

\bibliography{references}

% \appendix        % 切换到附录模式
% \setcounter{linenumber}{1}
\renewcommand{\thesection}{\arabic{section}} % 附录章节重新编号为 1,2,3...
\setcounter{section}{0} % 章节计数器清零
\renewcommand{\theequation}{A\arabic{equation}} % 公式编号形式：A1, A2, ...
\setcounter{equation}{0} % 公式编号重新从0开始
\renewcommand{\tablename}{Supplementary Table}
\renewcommand{\thetable}{\arabic{table}}

\clearpage       % 正文后强制换页
\section*{Appendix}  % 如果你想加一个总标Preprin题“附录”

\section{Theoretical Analysis}
\subsection{Retrieval in the Gradient Space}
We begin by considering the simplest case, an artificial neural network (ANN) consisting of a single linear layer. Let $X \in \mathbb{R}^{n \times k}$ denote the input matrix to this layer with batch size $k$, $Y \in \mathbb{R}^{m \times k}$ the output matrix, and $W \in \mathbb{R}^{m \times n}$ the corresponding weight matrix. The neural network employs the mean squared error (MSE) as the loss function:
\begin{equation}
    \mathcal{L} = \frac{1}{2} \| WX - Y \|_F^2 .
\end{equation}
Accordingly, the gradient of the loss function with respect to the weights $W$ can be expressed as
\begin{equation}
    \nabla_W \mathcal{L} = (WX - Y) X^T = \Delta X^T ,
\end{equation}
where $\Delta \in \mathbb{R}^{m \times k}$ is the error matrix. Therefore, the row vectors of the gradient matrix $\nabla_W \mathcal{L}$ lie within the vector space spanned by the column vectors of $X$.

\paragraph{The Gradient Updates Lie in the Span of Inputs}
We now extend this result to the case of a multi-layer fully connected neural network, where each layer is equipped with a nonlinear activation function. Consider an $L$-layer neural network with input $X_0 \in \mathbb{R}^{d_0 \times k}$, and for each layer $l=1,\dots,L$ define
\begin{equation}
Z_l = W_l X_{l-1}, \quad X_l = f_l(Z_l),
\label{eq:XtoZ}
\end{equation}
where $W_l \in \mathbb{R}^{d_l \times d_{l-1}}$, $b_l \in \mathbb{R}^{d_l}$, and $f_l$ is an elementwise activation function. The final output is $X_L$, and the mean squared error loss is
\begin{equation}
\mathcal{L} = \frac{1}{2} \|X_L - Y\|_F^2.
\label{eq:mseloss}
\end{equation}

We prove by backward induction that for any layer $l$, the gradient of the loss function with respect to $W_l$ admits the factorization
\begin{equation}
\nabla_{W_l} \mathcal{L} = \Delta_l X_{l-1}^T,
\label{eq:factorization}
\end{equation}
where $\Delta_l \in \mathbb{R}^{d_l \times k}$ is the backpropagated error matrix at layer $l$. 

For the output layer $l=L$, let $E = X_L - Y$. Then
\begin{equation}
\nabla_{W_L} \mathcal{L} = (X_L - Y) X_{L-1}^T = E X_{L-1}^T,
\end{equation}
so the claim holds with $\Delta_L = E$. Suppose it holds for some $l+1 \leq L$, \textit{i.e.}\ $\nabla_{W_{l+1}} \mathcal{L} = \Delta_{l+1} X_l^T$. By the chain rule of backpropagation, the error at layer $l$ is
\begin{equation}
\Delta_l
= \frac{\partial \mathcal{L}}{\partial Z_l}
= \frac{\partial \mathcal{L}}{\partial Z_{l+1}}\,\frac{\partial Z_{l+1}}{\partial X_l}\,\frac{\partial X_l}{\partial Z_l}
= (W_{l+1}^T \Delta_{l+1}) \circ f_l'(Z_l),
\end{equation}
where $\circ$ denotes elementwise multiplication. By induction, the factorization \ref{eq:factorization} holds for all layers $l=1,\dots,L$.

Therefore, the row vectors of $\nabla_{W_l} \mathcal{L}$ lie within the vector space spanned by the column vectors of $X_{l-1}$, \textit{i.e.},
\begin{equation}
\operatorname{Row}(\nabla_{W_l} \mathcal{L}) \subseteq \operatorname{Col}(X_{l-1}).
\end{equation}

\paragraph{Memory Retrieval to Avoid Forgetting}
We now consider the case where the model has already learned several batches of data. When learning new data, enforcing the gradients to remain orthogonal to the subspace spanned by the previously learned data ensures that the updated model yields unchanged loss when the old data are used as input.

In this case, for each layer $t$ we maintain a memory matrix $M_l \in \mathbb{R}^{d_{l-1}\times n_l}$ whose columns form an orthonormal basis of a subspace that contains the column space of the old input $X_{l-1}$, \textit{i.e.}, $\operatorname{Col}(X_{l-1}) \subseteq \operatorname{Col}(M_l)$. Let the projection matrix onto $\operatorname{Col}(M_l)$ be $P_l = M_l M_l^T$. We then decompose the gradient into a component within this subspace and an orthogonal component. Define
\begin{equation}
G_{\perp} = \nabla_{W_l}\mathcal{L}' (I - P_l),
\end{equation}
where $\nabla_{W_l} \mathcal{L}'$ denotes the gradient computed on the new data. 

Below we prove that using $G_{\perp}$ as the update direction (\textit{i.e.}\ $W_l \leftarrow W_l - \eta\, G_{\perp}$) does not change the loss on old data. Compute the action of $G_{\perp}$ on the forward inputs:
\begin{equation}
G_{\perp} X_{l-1}
= \big(\nabla_{W_l}\mathcal{L}\,(I-P_l)\big) X_{l-1}
= \nabla_{W_l}\mathcal{L} \big((I-P_l) X_{l-1}\big).
\end{equation}
Because $\operatorname{Col}(X_{l-1}) \subseteq \operatorname{Col}(M_l)$ we have $P_l X_{l-1} = X_{l-1}$, and therefore $(I-P_l)X_{l-1}=0$. Hence $G_{\perp} X_{l-1} = 0.$

Let $\Delta W = -\eta\,G_{\perp}$ be the parameter update applied to $W_l$, where $\eta$ is the learning rate. Its effect on the pre-activation at layer $l$ is
\begin{equation}
Z_l' = (W_l + \Delta W) X_{l-1} + b_l = W_l X_{l-1} + b_l + (\Delta W) X_{l-1} = Z_l + (\Delta W) X_{l-1}.
\end{equation}
Since $(\Delta W) X_{l-1} = -\eta\, G_{\perp} X_{l-1} = 0$, we obtain $Z_l' = Z_l$. Thus the activation $X_l = f_l(Z_l)$ and all subsequent layer inputs and outputs remain exactly unchanged. Therefore the final loss (\ref{eq:mseloss}) is exactly unchanged by the update, which means
\begin{equation}
\Delta \mathcal{L}=0
\label{eq:loss0}
\end{equation}

This completes the proof that projecting the current batch gradient to be orthogonal to the old gradient subspace guarantees exact invariance of the loss on the old data, thereby preserving the previously learned knowledge. Although we prove it at fully connected layers, the same principles apply to convolutional layers with appropriate structural adjustments.

\subsection{Consolidation and Forgetting}

To avoid storing all columns of $X_{l-1}$ (which quickly leads to full-rank memory and over-constraining), we perform a selective consolidation of the layer-wise memory via the SVD of $X_{l-1}$. For each layer $t$ let the matrix of inputs $X_{l-1}$ be 
\begin{equation}
X_{l-1} = \sum_{i=1}^{r_l} \sigma_{i,l}\, \mathbf{u}_{i,l} \mathbf{v}_{i,l}^T, 
\end{equation}
where $r_l=\operatorname{rank}(X_{l-1})$ and the singular values are in descending order $\sigma_{1,l} \ge \sigma_{2,l} \ge \cdots \ge \sigma_{r_l,l}\ge 0$. 
We choose a truncation level $r_l'\le r_l$ according to a retained-energy threshold $c_l\in[0,1]$:
\begin{equation}
r_l' = \arg\max_{r}\left( \frac{\sum_{i=1}^r \sigma_{i,l}^2}{\sum_{i=1}^{r_l} \sigma_{i,l}^2} \le c_l \right).
\end{equation}

The consolidated memory at layer $l$ is the matrix of left singular vectors, and the orthogonal projector onto the retained subspace.
\begin{equation}
M_l = \big[ \mathbf{u}_{1,l}, \mathbf{u}_{2,l}, \dots, \mathbf{u}_{r_l',l} \big],
\qquad P_l = M_l M_l^T
\end{equation}

The residual (forgotten) component of $X_{l-1}$ is
\begin{equation}
R_l := (I - P_l) X_{l-1}
= \sum_{i=r_l'+1}^{r_l} \sigma_{i,l}\, \mathbf{u}_{i,l} \mathbf{v}_{i,l}^T .
\end{equation}
Whose Frobenius norm satisfies
\begin{equation}
\|R_l\|_F^2 = \sum_{i=r_l'+1}^{r_l} \sigma_{i,l}^2
\approx \big(1 - c_l\big)\, \|X_{l-1}\|_F^2.
\label{eq:residual_energy}
\end{equation}

\paragraph{Single-layer Loss Upper Bound}
Consider an update at layer $l$ using the projected gradient
\begin{equation}
G_{\perp}^{(l)} = \nabla_{W_l} \mathcal{L}' \,(I - P_l),
\qquad \Delta W_l = -\eta\, G_{\perp}^{(l)},
\end{equation}
where $\nabla_{W_l}\mathcal{L}'$ is the gradient computed on the new data. The change of the pre-activation for the old inputs equals
\begin{equation}
\Delta Z_l = (\Delta W_l) X_{l-1} = -\eta\, \nabla_{W_l} \mathcal{L}' \, R_l.
\end{equation}

Using the first-order approximation for the old-data loss (\ref{eq:mseloss}) and the Frobenius inner product we obtain the local loss perturbation bound. By the Cauchy-Schwarz inequality, we have
\begin{align}
|\Delta \mathcal{L}_l|
&\approx \big| \langle \nabla_{W_l} \mathcal{L} \,,\, \Delta W_l \rangle_F \big|
= \eta \big| \langle \nabla_{W_l} \mathcal{L} \,,\, \nabla_{W_l} \mathcal{L}' (I-P_l) \rangle_F \big| \nonumber\\
&\le \eta \,\| \nabla_{W_l} \mathcal{L} \|_F \, \| \nabla_{W_l} \mathcal{L}' \|_F \, \| R_l \|_F.
\label{eq:single_layer_bound}
\end{align}

\paragraph{Multi-layer Loss Upper Bound}
To account for nonlinear activations and propagation through subsequent layers, assume each activation $f_l$ is $K_l$-Lipschitz in the sense that for the pre-activation perturbation $\Delta Z_l$,
\begin{equation}
\|\Delta X_l\|_F = \|f_l(Z_l+\Delta Z_l)-f_l(Z_l)\|_F \le K_l \|\Delta Z_l\|_F,
\label{eq:lip}
\end{equation}
where for elementwise activations one can take $K_l = \sup_z |f_l'(z)|$ when this supremum exists.

For layer $l$, we compute the error propagated from the input to the next layer. According to the formula (\ref{eq:XtoZ}), we have:
\begin{equation}
\Delta Z_{l+1} = W_{l+1}\, \Delta X_l.
\label{eq:nextZ}
\end{equation}
Taking the Frobenius norm and using the submultiplicativity \(\|AB\|_F \le \|A\|_2 \|B\|_F\) gives
\begin{equation}
\|\Delta Z_{l+1}\|_F \le \|W_{l+1}\|_2 \,\| \Delta X_l \|_F.
\label{eq:nextZnorm}
\end{equation}
Then applying the activation Lipschitz bound \eqref{eq:lip} at layer \(l+1\),
\begin{equation}
\|\Delta X_{l+1}\|_F
\le K_{l+1} \,\| \Delta Z_{l+1} \|_F
\le K_{l+1} \,\|W_{l+1}\|_2 \,\| \Delta X_l \|_F.
\label{eq:one_step}
\end{equation}

The source pre-activation perturbation $\Delta Z_l$ at layer \(l\) caused by the parameter update \(\Delta W_l\) is
\begin{equation}
\Delta Z_l = (\Delta W_l) X_{l-1}.
\end{equation}
In particular, when \(\Delta W_l = -\eta\,\nabla_{W_l} \mathcal{L}' (I-P_l)\) we have the norm bound
\begin{equation}
\| \Delta Z_l \|_F \le \eta \,\| \nabla_{W_l} \mathcal{L}' \|_F \, \| R_l \|_F,
\label{eq:DeltaZ_bound}
\end{equation}

The perturbation at layer $l$ induces a propagated perturbation at deeper layers. By combining Equations \ref{eq:one_step} and \ref{eq:DeltaZ_bound}, we obtain the perturbation bound of the final-layer activation
\begin{equation}
\|\Delta X_L\|_F
\le \sum_{l=1}^L \Bigg( \prod_{k=l+1}^L K_k \,\|W_k\|_2 \Bigg) \, K_l \,\eta \,\| \nabla_{W_l} \mathcal{L}' \|_F \, \| R_l \|_F,
\label{eq:propagated_error}
\end{equation}
where $\|\cdot\|_2$ denotes the spectral norm and the product for $l=L$ is interpreted as $1$.

Now we define the Jacobian from the pre-activation at layer \(l\) to the final activation \(X_L\) by
\begin{equation}
J_{L,l} \;=\; \frac{\partial X_L}{\partial X_l} \;=\; \frac{\partial X_L}{\partial X_{L-1}}\frac{\partial X_{L-1}}{\partial X_{L-2}} \dots \frac{\partial X_{l+1}}{\partial X_{l}}.
\end{equation}
Writing the layerwise linearization explicitly (for elementwise activations \(f_k\)),
\begin{equation}
J_{L,l} \;=\; \Big( D_L W_L \Big)\Big( D_{L-1} W_{L-1} \Big)\cdots \Big( D_{l+1} W_{l+1} \Big)\,,
\end{equation}
where \(D_k=\operatorname{diag}\big(f_k'(Z_k)\big)\). By submultiplicativity of the spectral norm and using \(\|D_k\|_2 \le K_k\) we obtain the standard upper bound
\begin{equation}
\label{eq:jacobian_prod_bound}
\|J_{L,l}\|_2 \;=\; \sigma_{\max}(J_{L,l})
\le \prod_{k=l+1}^L \big( K_k \,\|W_k\|_2 \big).
\end{equation}

Therefore, we obtain a tighter upper bound for Equation \ref{eq:propagated_error}:
\begin{equation}
\|\Delta X_L\|_F
\le \sum_{l=1}^L \sigma_{\max}(J_{L,l}) K_l \,\eta \,\| \nabla_{W_l} \mathcal{L}' \|_F \, \| R_l \|_F,
\label{eq:propagated_error_J}
\end{equation}

Finally, using the first-order relation between output perturbation and loss change for the MSE loss,
\begin{equation}
\Delta \mathcal{L} \approx \langle \nabla_{X_l} \mathcal{L} \,,\, \Delta X_l \rangle_F
\le \| \nabla_{X_l} \mathcal{L} \|_F \, \| \Delta X_l \|_F,
\end{equation}
and Substituting \eqref{eq:propagated_error_J} yields the global bound
\begin{equation}
\boxed{%
\Delta \mathcal{L} \le \eta \,\| \nabla_{X_l} \mathcal{L} \|_F \sum_{l=1}^L \sigma_{\max}(J_{L,l}) K_l \,\| \nabla_{W_l} \mathcal{L}' \|_F \, \| R_l \|_F .
}
\label{eq:final_bound}
\end{equation}

\paragraph{Remarks}
\begin{itemize}
  \item When $c_l = 1$ for every layer $l$ (\textit{i.e.},\ no truncation), the residual $\|R_l\|_F=0$ and the RHS of \eqref{eq:final_bound} vanishes, recovering the exact invariance result (\ref{eq:loss0}).
  \item In practical applications, the value of $c_l$ is very close to 1, which makes $\|R_l\|_F$ (\ref{eq:residual_energy}) very small. Moreover, when the old data has been sufficiently trained to convergence, the value of $\|\nabla_{X_l} \mathcal{L}\|_F$ is also close to 0. In addition, the learning rate $\eta$ is small. Therefore, this consolidation process can largely preserve old knowledge and mitigate forgetting.
  \item This conclusion also holds for the forgetting process, which is analogous to consolidation. Although the model discards memories that contribute the least to 'predicting the future', it simultaneously minimizes forgetting of prior knowledge, thus balancing plasticity and stability in LL.
\end{itemize}

% \subsection{Implementation of the Core Memory Encoding}
% 附录1.3

\section{Implementation Details}

\subsection{Algorithm for Lifelong Learning with AdaMLearn}

To implement lifelong learning with adaptive engram memorization, we design a procedure that integrates retrieval, encoding, consolidation and forgetting into a single learning loop. As Algorithm~\ref{alg:engram} illustrates, the model learns sequential tasks and selectively preserving knowledge representations through adaptive memory updating. This enables efficient memory reuse and stable performance across tasks.

\begin{algorithm}[ht]
\caption{Lifelong Learning with Adaptive Engram Memorization}
\label{alg:engram}
\begin{algorithmic}[1]
\Require model $f(\cdot)$, model parameters $\theta$, tasks $\{D_\tau\}_{\tau=1}^T$, projection memory $\mathcal{M}=\{M^{\left(l\right)}\}_{l=1}^L$, consolidation thresholds $\{c_\tau^{(l)}\}$, forgetting thresholds $\{\beta_\tau^{(l)}\}$.
\Ensure updated model $\theta$ and memory $\mathcal{M}$

\For{$\tau = 1$ \textbf{to} $T$}
    \Repeat
        \State Sample mini-batch $B \sim D_\tau$, and compute gradients $\nabla_{\theta} \mathcal{L}(B; \theta)$
        \For{each layer $l = 1,\dots,L$}
            \If{$M^{\left(l\right)}$ is not empty}
                \State \textbf{Memory retrieval (projection):}
                \State $\nabla_{\theta^{(l)}} \mathcal{L} \gets \nabla_{\theta^{(l)}} \mathcal{L} - Project(\nabla_{\theta^{(l)}} \mathcal{L})$ // eq (\ref{eq:R1})
            \EndIf
        \EndFor
        \State Update parameters: $\theta \gets \theta - \alpha \nabla_{\theta}\mathcal{L}$ // learning rate $\alpha$
    \Until{convergence or stopping criterion}
    \State 
    \State \textbf{Memory encoding:}
    \State  $S_\tau \gets Selection(D_\tau, f_\theta(\cdot))$ // eq (\ref{eq:E3})
    \State  $\{X_\tau^{(l-1)}\} \gets HookLayerwiseInputs(f_\theta(S_\tau))$ for layer $l=1,\dots,L$
    \State 
    \State \textbf{Adaptive memory consolidation \& forgetting:}
    \For{each layer $l=1,\dots,L$}
        \State $(U, \Sigma, \_) \gets SVD(X_\tau^{(l-1)})$ // eq (\ref{eq:CF1})
        \State $U' \gets RankSelect1(U, \Sigma, c_\tau^{(l)})$ // eq (\ref{eq:CF2})
        \If{$\tau = 1$}
            \State $M^{\left(l\right)} \gets U'$ // \textbf{Only consolidation}
        \Else
            \State $(\hat X_\tau^{(l-1)}, \tilde X_\tau^{(l-1)}) \gets ProjectSplit(X_\tau^{(l-1)}, M^{\left(l\right)})$ // eq (\ref{eq:CF3})
            \State $(\hat U, \hat \Sigma, \_) \gets SVD(\hat X_\tau^{(l-1)})$, $(\tilde U, \tilde \Sigma, \_) \gets SVD(\tilde X_\tau^{(l-1)})$ // eq (\ref{eq:CF4})
            \State $\tilde U' \gets RankSelect2(\tilde U, \tilde \Sigma, \hat \Sigma, c_\tau^{(l)})$ // eq (\ref{eq:CF5}) \textbf{consolidation}
            \State $\hat U' \gets RankSelect1(\hat U, \hat \Sigma, \beta_\tau^{(l)})$ // eq (\ref{eq:CF6}) \textbf{forgetting}
            \State $M^{\left(l\right)} \gets Concat(\hat U', \tilde U')$ // eq (\ref{eq:CF7})
        \EndIf
    \EndFor
\EndFor

\State \Return $\theta,\ \mathcal{M}$
\end{algorithmic}
\end{algorithm}

\subsection{Neural Network Architecture}
Following the prior work~\cite{GPM}, we employ different backbones according to dataset scale and complexity. Specifically, AlexNet is used for CIFAR-100, LeNet for CIFAR-superclass, ResNet18(32×32) for the FIVE benchmark, and ResNet18(84×84) for Mini-ImageNet. 

All neural networks follow standard configurations with task-incremental learning. The classifier heads are defined separately for each task, and the shared feature extractor maintains identical structure across learners. The detailed architectures are summarized in Supplementary Tables 1-5, where the convolutional and residual blocks are explicitly described. 

\begin{table}[htbp]
\centering
\caption{Neural network architecture (AlexNet) for CIFAR-100.}
\label{tab:alexnet}
\begin{tabular}{lccccc}
\toprule
Layer & Channels & Kernel & Stride & Padding & Dropout \\
\midrule
Input & $3$ & - & - & - & - \\

Conv1 + BN & $128$ & $4 \times 4$ & $1$ & - & 0.2 \\
MaxPool1 & $128$ & $2 \times 2$ & $2$ & - & - \\

Conv2 + BN & $256$ & $3 \times 3$ & $1$ & - & 0.2 \\
MaxPool2 & $256$ & $2 \times 2$ & $2$ & - & - \\

Conv3 + BN & $512$ & $3 \times 3$ & $1$ & - & 0.5 \\
MaxPool3 & $512$ & $2 \times 2$ & $2$ & - & - \\

Flatten & $2048$ & - & - & - & - \\

FC1 + BN & $4096$ & - & - & - & 0.5 \\
FC2 + BN & $4096$ & - & - & - & 0.5 \\

Classifier ($\times 10$ heads) & $10$ & - & - & - & - \\
\bottomrule
\end{tabular}
\end{table}

\begin{table}[htbp]
\centering
\caption{Neural network architecture (LeNet) for CIFAR-superclass.}
\label{tab:lenet}
\begin{tabular}{lccccc}
\toprule
Layer & Channels & Kernel & Stride & Padding & Dropout \\
\midrule
Input & $3$ & - & - & - & - \\

Conv1 + ReLU + LRN & $40$ & $5 \times 5$ & $1$ & $2$ & - \\
MaxPool1 & $40$ & $3 \times 3$ & $2$ & $1$ & - \\

Conv2 + ReLU + LRN & $100$ & $5 \times 5$ & $1$ & $2$ & - \\
MaxPool2 & $100$ & $3 \times 3$ & $2$ & $1$ & - \\

Flatten & $6400$ & - & - & - & - \\

FC1 & $1600$ & - & - & - & - \\
FC2 & $800$ & - & - & - & - \\

Classifier ($\times 20$ heads) & $5$ & - & - & - & - \\
\bottomrule
\end{tabular}
\end{table}

\begin{table}[h]
\centering
\caption{Structure of BasicBlock used in ResNet18.}
\label{tab:basicblock}
\begin{tabular}{lccccc}
\toprule
Layer & Channels & Kernel & Stride & Padding & Dropout \\
\midrule
Conv1 + BN + ReLU & $C_{in} \rightarrow C_{out}$ & $3\times3$ & $s$ & 1 & - \\
Conv2 + BN & $C_{out} \rightarrow C_{out}$ & $3\times3$ & 1 & 1 & - \\
Shortcut & $C_{in} \rightarrow C_{out}$ & $1\times1$ (if needed) & $s$ & 0 & - \\
ReLU (post-addition) & - & - & - & - & - \\
\bottomrule
\end{tabular}
\end{table}

\begin{table}[h]
\centering
\caption{Neural network architecture (ResNet18(32×32)) for FIVE benchmark.}
\label{tab:resnet32x32}
\begin{tabular}{lccccc}
\toprule
Layer & Channels & Kernel & Stride & Padding & Dropout \\
\midrule
Input & $C$ & - & - & - & - \\
Conv1 + BN + ReLU & 64 & $3\times3$ & 1 & 1 & - \\
\midrule
\multicolumn{6}{c}{\textbf{Stage 1: BasicBlock × 2}} \\
\midrule
Block1 & 64 & $3\times3$ & 1 & 1 & - \\
Block2 & 64 & $3\times3$ & 1 & 1 & - \\
\midrule
\multicolumn{6}{c}{\textbf{Stage 2: BasicBlock × 2}} \\
\midrule
Block1 & 128 & $3\times3$ & 2 & 1 & - \\
Block2 & 128 & $3\times3$ & 1 & 1 & - \\
\midrule
\multicolumn{6}{c}{\textbf{Stage 3: BasicBlock × 2}} \\
\midrule
Block1 & 256 & $3\times3$ & 2 & 1 & - \\
Block2 & 256 & $3\times3$ & 1 & 1 & - \\
\midrule
\multicolumn{6}{c}{\textbf{Stage 4: BasicBlock × 2}} \\
\midrule
Block1 & 512 & $3\times3$ & 2 & 1 & - \\
Block2 & 512 & $3\times3$ & 1 & 1 & - \\
\midrule
AvgPool2d + Flatten & 2048 & - & - & - & - \\
Classifier (×5 heads) & 10 & Linear & - & - & - \\
\bottomrule
\end{tabular}
\end{table}

\begin{table}[h]
\centering
\caption{Neural network architecture (ResNet18(84×84)) for Mini-imagenet.}
\label{tab:resnet84x84}
\begin{tabular}{lccccc}
\toprule
Layer & Channels & Kernel & Stride & Padding & Dropout \\
\midrule
Input & $3$ & - & - & - & - \\
Conv1 + BN + ReLU & 64 & $3\times3$ & 1 & 1 & - \\
\midrule
\multicolumn{6}{c}{\textbf{Stage 1: BasicBlock × 2}} \\
\midrule
Block1 & 64 & $3\times3$ & 1 & 1 & - \\
Block2 & 64 & $3\times3$ & 1 & 1 & - \\
\midrule
\multicolumn{6}{c}{\textbf{Stage 2: BasicBlock × 2}} \\
\midrule
Block1 & 128 & $3\times3$ & 2 & 1 & - \\
Block2 & 128 & $3\times3$ & 1 & 1 & - \\
\midrule
\multicolumn{6}{c}{\textbf{Stage 3: BasicBlock × 2}} \\
\midrule
Block1 & 256 & $3\times3$ & 2 & 1 & - \\
Block2 & 256 & $3\times3$ & 1 & 1 & - \\
\midrule
\multicolumn{6}{c}{\textbf{Stage 4: BasicBlock × 2}} \\
\midrule
Block1 & 512 & $3\times3$ & 2 & 1 & - \\
Block2 & 512 & $3\times3$ & 1 & 1 & - \\
\midrule
AvgPool2d + Flatten & 12800 & - & - & - & - \\
Classifier (×20 heads) & 5 & Linear & - & - & - \\
\bottomrule
\end{tabular}
\end{table}

\subsection{Training Regime and Hyperparameter}
This subsection reports details in the implementation of the LL methods. 
For AdaMLearn, we trained our neural network using SGD optimizer with the learning rate of $0.01$ for 50 epochs or $patience = 6$. The batch size is defaulted to 32 or 64.
%For other comparison methods, according to the recommended configurations of the Mammoth benchmark, we used ResNet18 as the backbone for all experiments. 
For each task, we trained our Neural Network using SGD optimizer with the learning rate of $0.03$ for 50 epochs. The batch size is defaulted to 32. For every experiment, we chose seeds from $0$ to $4$ for five runs, respectively, and reported the average results for comparison.
Except for Mini-ImageNet, the other three experiments were conducted under a fixed task order, whereas the task order in Mini-ImageNet varied with different random seeds.

\begin{itemize}
    \item AdaMLearn: We set different parameters according to the characteristics of each dataset. For CIFAR-100, $c_T = 0.97 + 0.003 \times T$(for T = 0 to 9) and $\beta_T = 0.99$; for CIFAR-superclass, $c_T = 0.98 + 0.001 \times T$(for T = 0 to 19) and $\beta_T = 0.9992$; for FIVE, $c_T = 0.98$ and $\beta_T = 0.9999$; and for Mini-ImageNet, $c_T = 0.99$ and $\beta_T = 0.99996$. The parameters are not superscripted with $l$ because they are shared across all neural network layers. The selection ratio for the core set is 0.01.
    \item Elastic Weight Consolidation (EWC~\cite{2017ewc}): We set $\lambda_{EWC}=25.0$ and $\gamma=1.0$.
    \item Learning without Forgetting (LwF~\cite{2017LwF}): We set the penalty weight $\lambda_o$ to $0.05$ and the softmax temperature $T$ to $20$.
    \item Synaptic Intelligence (SI~\cite{2017si}): The surrogate loss weight $c$ was set to $15$, and the additional damping parameter $\xi$ was set to $1$.
    \item Averaged Gradient Episodic Memory (AGEM~\cite{2018agem}): The buffer size was set to $500$.
    \item Experience Replay (ER~\cite{2019er}): Similarly, we chose the buffer size $500$ for experiments.
    \item Dark Experience Replay (DER~\cite{2020der}): We chose the penalty weight $\alpha=3$ and buffer size $500$. 
    \item Learning a Unified Classifier Incrementally via Rebalancing (LUCIR~\cite{2019lucir}): We set regularization weight for embedding cosine similarity $\lambda_{base}$ to $15$. For margin-ranking loss, the margin $m$ is set to $0.5$, $K$ was set to $2$, and its regularization weight $\lambda_{mr}$ was set to $10$. Number of finetuning epochs on coreset was set to $20$, with learning rate $0.01$. 
    \item Bias Correction (BiC~\cite{2023bic}): We set the softmax temperature $T=20$. Specifically, we only adopted the L2 regularization with weight $1e-5$ when training on CIFAR-superclass.
    \item eXtended Dark Experience Replay (X-DER~\cite{2022xder}): We set $\alpha=0.3$, $\beta = 0.8$, $\gamma = 0.85$, and $\lambda = 0.05$.
\end{itemize}

% \section{Additional Experimental Analyses}
% 附录3

\end{document}